%% file: sn-article.tex
\documentclass[pdflatex,sn-mathphys-num]{sn-jnl}

\usepackage{graphicx}%
\usepackage{multirow}%
\usepackage{amsmath,amssymb,amsfonts}%
\usepackage{amsthm}%
\usepackage{mathrsfs}%
\usepackage[title]{appendix}%
\usepackage{xcolor}%
\usepackage{textcomp}%
\usepackage{manyfoot}%
\usepackage{booktabs}%
\usepackage{algorithm}%
\usepackage{listings}%

\usepackage{algorithmic}%
\usepackage{subfigure}
\usepackage{soul}
\usepackage{makecell}

\theoremstyle{thmstyleone}%
\theoremstyle{thmstyletwo}%

\theoremstyle{thmstylethree}%

\begin{document}

\title[Article Title]{\(X\)-evolve: Solution space evolution powered by large language models}


\author[1]{\fnm{Yi} \sur{Zhai}}\email{zhaiyi0@mail.ustc.edu.cn}
\author[1]{\fnm{Zhiqiang} \sur{Wei}}\email{weizhiqiang@mail.ustc.edu.cn}
\author[1]{\fnm{Ruohan} \sur{Li}}\email{lrh2947528032@mail.ustc.edu.cn}

\author[2]{\fnm{Keyu} \sur{Pan}}\email{pankeyu96@gmail.com}
\author[1]{\fnm{Shuo} \sur{Liu}}\email{zkdliushuo@mail.ustc.edu.cn}
\author[3]{\fnm{Lu} \sur{Zhang}}\email{luzha@ustc.edu.cn}

\author[1]{\fnm{Jianmin} \sur{Ji}}\email{jianmin@ustc.edu.cn}
\author[1]{\fnm{Wuyang} \sur{Zhang}}\email{wuyangz@ustc.edu.cn}
\author[1]{\fnm{Yu} \sur{Zhang}}\email{yuzhang@ustc.edu.cn}
\author*[1]{\fnm{Yanyong} \sur{Zhang}}\email{yanyongz@ustc.edu.cn}






\affil[1]{\orgname{University of Science and Technology of China}, \orgaddress{ \country{China}}}

\affil[2]{\orgname{ByteDance Ltd.}, \orgaddress{ \country{China}}}

\affil[3]{\orgdiv{Institute of Artificial Intelligence}, \orgname{Hefei Comprehensive National Science Center}, \orgaddress{ \country{China}}}

\input{0_abstract}

\maketitle

\input{1_intro}

\input{2_method}
\input{3_results}
\input{4_ablations}

\input{5_relate}
\input{6_discussion}
\input{7_acknowledgements}

\begin{appendices}

\input{a_discovered_program}
\input{a_bin_packing}

\input{a_algorithm}

\input{a_hyperparameters}

\end{appendices}

\newpage
\bibliography{sn-bibliography}

\end{document}

%% file: 0_abstract.tex
\abstract{

While combining large language models (LLMs) with evolutionary algorithms (EAs) shows promise for solving complex optimization problems, current approaches typically evolve individual solutions, often incurring high LLM call costs. We introduce \(X\)-evolve, a paradigm-shifting method that instead evolves solution spaces \(X\) (sets of individual solutions) — subsets of the overall search space \(S\). In \(X\)-evolve, LLMs generate tunable programs wherein certain code snippets, designated as parameters, define a tunable solution space. A score-based search algorithm then efficiently explores this parametrically defined space, guided by feedback from objective function scores. This strategy enables broader and more efficient exploration, which can potentially accelerate convergence at a much lower search cost, requiring up to two orders of magnitude fewer LLM calls than prior leading methods. We demonstrate \(X\)-evolve’s efficacy across three distinct hard optimization problems. For the cap set problem, we discover a larger partial admissible set, establishing a new tighter asymptotic lower bound for the cap set constant (\(C \ge 2.2203\)). In information theory, we uncover a larger independent set for the 15-vertex cycle graph (\(\mathcal{C}_{15}^{\boxtimes 5}\), size 19,946), thereby raising the known lower bound on its Shannon capacity. Furthermore, for the NP-hard online bin packing problem, we generate heuristics that consistently outperform standard strategies across established benchmarks. By evolving solution spaces, our method considerably improves search effectiveness, making it possible to tackle high-dimensional problems that were previously computationally prohibitive.

}

\keywords{Large language model, Evolutionary algorithm, Optimization problem, Cap set, Shannon capacity of cycle graph}

%% file: 1_intro.tex
\section{Introduction}\label{intro}

With the rapid advancement of large language models (LLMs), 
many recent studies~\cite{romera2024mathematical, alpha2025alexander, lehman2023evolution, hemberg2024evolving, shojaee2024llm, liu2024evolution, fernando2023promptbreeder, chen2023evoprompting} have explored combining LLMs with evolutionary algorithms (EAs)~\cite{goldberg1989genetic, koza1994genetic, banzhaf1998genetic} to tackle complex optimization problems~\cite{nocedal1999numerical, tao2006additive, shannon1956zero, golub2013matrix, conway2013sphere}.
Generally, optimization problems can be defined as follows: Given an objective function \(f: S \rightarrow \mathbb{R}\) defined over a feasible domain \(S\), the goal is to find an optimal solution \(x^* \in S\) such that: \(f(x^*) = \max_{x \in S} f(x)\). The objective functions discussed in this paper are all evaluable (i.e., explicitly computable or measurable), providing an evaluation score for each input \(x\).
Minimization problems can often be transformed into maximization problems by negating the objective function.
In this paper, we also adopt the LLM+EA approach for solving optimization problems. However, unlike previous work primarily focused on evolving individual solutions \(x\), we propose an alternative approach, named \textbf{\(X\)-evolve}, that \textbf{evolve}s solution spaces \textbf{\(X\)} (sets of individual solutions) — subsets of the search space \(S\)\footnote{For conciseness, \(S\) denotes both the mathematically defined feasible domain and the search space in optimization; while not always strictly identical, they are closely related.}. This strategy enables a more comprehensive exploration of the search space, potentially accelerating convergence and reducing the overall search cost.

LLMs are generally not well-suited for direct application to complex optimization problems due to a few key factors. Firstly, regarding training data, many complex optimization problems remain open, with their optimal solutions in real-world contexts still under exploration. This scarcity of optimal solution data for training severely limits the ability of LLMs to directly generate high-quality solutions. Secondly, concerning model characteristics and fine-tuning challenges, LLMs are inherently general-purpose. Fine-tuning them for a specific optimization problem using tailored evaluation scores often proves impractical due to prohibitive computational costs and extensive data requirements~\cite{kaplan2020scaling, hu2022lora}. Consequently, LLMs typically lack the explicit, task-specific scoring feedback essential for precise optimization procedures like gradient descent. In fact, current primary methods for enhancing LLM capabilities and aligning them with human preferences, such as reinforcement learning from human feedback (RLHF)~\cite{ouyang2022training, christiano2017deep, stiennon2020learning, schulman2017proximal, rafailov2023direct}, rely on preference data (e.g., choose-rejected pairs~\cite{bradley1952rank, burges2005learning, ziegler2019fine} from curated tasks). This training aims to improve conversational style, generalization, and instruction following, which differs significantly from the targeted training and specific scoring feedback required to tackle complex optimization problems.

One important reason for using evolutionary algorithms~\cite{goldberg1989genetic, schwefel1993evolution, eiben2015introduction}, a form of heuristic search~\cite{russell2016artificial}, to solve optimization problems is that the objective function may be non-differentiable. For example, if the score involves non-differentiable operators~\cite{nocedal1999numerical} like argmax or relies on black-box evaluations~\cite{rios2013derivative}, gradient-based methods~\cite{nocedal1999numerical, rumelhart1986learning, robbins1951stochastic} become impractical. 
A key challenge in evolutionary algorithms is the explicit definition of the search space \(S\) that the algorithm explores within
~\cite{o2010open, romera2024mathematical, alpha2025alexander}. Each candidate solution must mutate within this space, requiring clear rules for allowable transformations. This often needs to be tailored to the structure of the specific problem and relies on expert knowledge. If the search space is too large, the computational cost can grow rapidly, making the search infeasible. Conversely, if the space is too narrow or excessively pruned, the search becomes less effective with high-quality solutions excluded.

While LLM and EA each face challenges in complex optimization problems, the marriage of the two offers unique advantages. Firstly, LLMs are ideal “evolution engines”. When using LLMs for evolution, a prompt typically includes two key components: reference implementations and an evolution instruction. The reference implementations provide the “inheritance” foundation, allowing the LLM to learn from existing strengths. The LLM’s generation process is essentially a probabilistic sampling process, inherently supporting “mutation”. This combination of “inheritance” and “mutation” is exactly the core of evolutionary algorithms. Moreover, unlike traditional evolutionary algorithms, LLM-driven evolution does not require explicitly defined rules for the search space -- a single evolution instruction is sufficient.

For open, complex optimization problems, LLMs often possess limited prior knowledge of optimal solutions due to scarce real-world information. Consequently, prompt understanding (including score-sampled reference implementations and instructions) becomes more critical. In other words, evolution is driven primarily by scoring feedback and instruction-following, rather than prior knowledge. While LLMs can quickly generate high-quality solutions for simple optimization problems, for complex ones, the evolution process is often lengthy and resource-intensive. To our knowledge, existing LLM+EA approaches typically evolve individual solutions directly, which can be inefficient and potentially costly (see \S\ref{ablation_summary} for detailed analysis).
This paper proposes evolving solution spaces, subsequently explored using a scoring feedback-based space search. Our approach aims to reduce overall search costs, thereby enabling its application to more computationally demanding tasks.

We draw inspiration from FunSearch~\cite{romera2024mathematical}, which uses LLMs to generate solution-generating programs rather than directly outputting the solutions themselves. This approach enhances interpretability, as the generated programs are inherently more understandable, thereby facilitating collaboration with domain experts. Given that each deterministic solution-generating program yields a unique solution, we consider the program and its output equivalent. Consequently, throughout this paper, the term “solution” encompasses both the program and the final result it produces.

\input{pic/introduction}

Fig.~\hyperref[pic_introduction]{\ref*{pic_introduction}(a)} shows the prompt template, which includes reference implementations, an evolution instruction, and a newly introduced tunable marker instruction, in \(X\)-evolve. The tunable marker identifies code snippets within the program that can be tuned, forming what we call “tunable programs” (Fig.~\hyperref[pic_introduction]{\ref*{pic_introduction}(b)}). Program analysis tools are used to extract tunable components, with each component defining a decision space \(D_i\). Sampling one decision from each decision space and replacing the tunable structure with these decisions yield an evaluable program. These evaluable programs collectively form the solution space \(X\) (Fig.~\hyperref[pic_introduction]{\ref*{pic_introduction}(c)}), where each solution represents a sample of the tunable program. The size of this solution space is the product of the sizes of all decision spaces, given by \(|X| = \prod_{i=1}^n |D_i|\). Thus, the optimization task is transformed into a sequential decision-making problem. This solution space is a subset of the larger search space \(S\).

To efficiently explore the solution space, we employ \(X\)-search, a score-based space search method. The process begins by randomly sampling decision spaces to create an initial batch of evaluable programs. Each program is then evaluated, and its overall score is propagated back to the individual decisions that formed it. If a decision appears in multiple programs, it retains the highest score it achieved. In subsequent rounds, we use this scoring information to guide the search. New programs are sampled using softmax selection, which favors decisions with higher scores. This iterative process of sampling, evaluation, and scoring continues until the top program score fails to improve for a set number of rounds. Finally, we compact the tunable program's structure by retaining only the top-\(K\) performing decisions or fixing any decision space with only one remaining option. This compacted tunable program (Fig.~\hyperref[pic_introduction]{\ref*{pic_introduction}(d)}) is then stored in a database, serving as a scored reference implementation for future use.

We demonstrate the efficacy of our solution space evolution method, \(X\)-evolve, by applying it to three distinct and challenging optimization problems. For the notoriously difficult cap set problem~\cite{grochow2019new, tao2006additive, tyrrell2022new, romera2024mathematical}, \(X\)-evolve establishes a new, tighter asymptotic lower bound for the cap set constant (\(C \ge 2.2203\)) by discovering a larger partial admissible set in a previously unexplored dimension \(\mathcal{A}(27,19)\). It also matches state-of-the-art results in constructing maximal cap sets for dimensions 3 through 8 with significantly improved search efficiency. In the domain of information theory, \(X\)-evolve successfully rediscovers the best-known maximum independent set sizes for the strong product powers of cycle graphs \(\mathcal{C}_9\), \(\mathcal{C}_{11}\), and \(\mathcal{C}_{13}\). Crucially, for the 15-vertex cycle graph (\(\mathcal{C}_{15}\)), our approach uncovers a larger independent set (size 19,946 for \(\mathcal{C}_{15}^{\boxtimes 5}\)), thereby improving the established lower bound~\cite{de2024asymptotic}. Finally, for the NP-hard online bin packing problem~\cite{coffman1984approximation}, heuristics generated by \(X\)-evolve consistently outperform standard strategies such as first-fit and best-fit across the widely-used OR-Library benchmarks~\cite{beasley1990or} and demonstrate robust performance on custom datasets (sampled from a Weibull distribution~\cite{castineiras2012weibull, angelopoulos2023online}) mimicking real-world scenarios.

Our method's efficacy stems from addressing two fundamental challenges: the lack of direct evaluative scoring feedback for LLMs and the difficulty of search space construction in EAs. We leverage LLMs not to produce singular solutions, but to generate broad solution spaces. This allows for the application of score-based feedback to guide exploration within these LLM-defined spaces, effectively providing the evaluative loop that LLMs typically miss. Furthermore, by using LLMs to define and iteratively refine these solution spaces, we circumvent the need for deep expert knowledge in search space design, thereby making complex problem-solving more accessible. Crucially, this solution space is not static but is iteratively improved by the LLM. This evolutionary nature ensures that the search is not limited by a single, potentially suboptimal initial generation from the LLM, allowing for progressive refinement towards higher-quality solutions. Consequently, our method enables the effective exploration of high-dimensional spaces, which were previously computationally prohibitive, paving the way for breakthroughs in challenging scientific problems.

%% file: pic/introduction.tex
\begin{figure}[t]%
    \centering
    \includegraphics[scale=0.4]{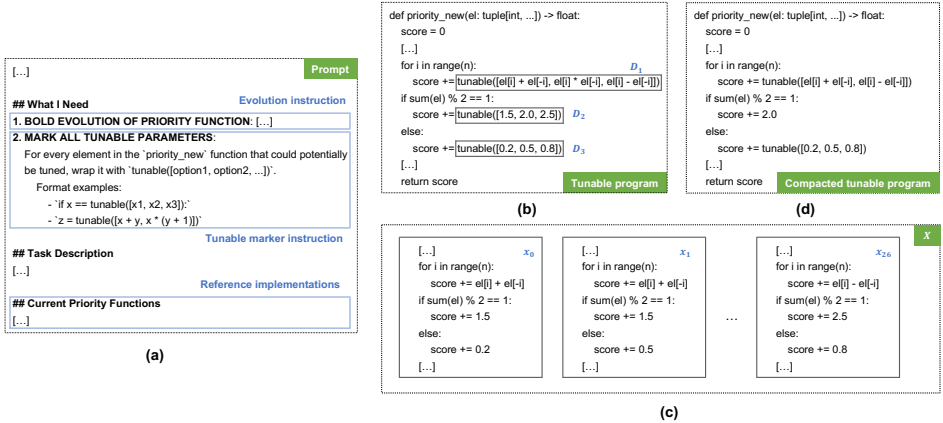} 

    \caption{
    \textbf{An \(X\)-evolve pipeline example.} [...] denotes omitted content for brevity.
\textbf{(a)} The LLM prompt, with three core parts: an evolutionary instruction, a tunable marker instruction, and reference implementations.
\textbf{(b)} An LLM-generated tunable program example, containing three tunable code snippets, each forming a decision space \(D\). Each decision space \(D\) has multiple decisions, with this example showing three decisions per decision space.
\textbf{(c)} The solution space \(X\), parsed from the tunable program. Its size is the product of the sizes of all decision spaces \(D\). This example has 27 solutions, each a sample of the tunable program.
\textbf{(d)} The compacted tunable program, which retains only tunable decisions found in the top-\(K\) highest-scoring programs evaluated in \textbf{c}. If a decision space has only one decision (e.g., \(D_2\)), its structure is removed, leaving just that decision.
}%
    \label{pic_introduction}%

\end{figure}

%% file: 2_method.tex
\section{\(X\)-evolve}\label{x_evolve}

\input{pic/overview_simple}

A high-level overview of \(X\)-evolve is presented in Fig.~\ref{pic_overview_simple}. The workflow for solving optimization problems using \(X\)-evolve proceeds as follows. Initially, the user provides the optimization problem to be solved and an evaluation function. This function assesses the quality (i.e., score) of a solution for the given optimization problem. Our framework then models this user-provided optimization problem as a search problem, which is addressed by combining a greedy search with an optimizable priority function. Subsequently, the core task is to search for this priority function. \(X\)-evolve features two primary components: solution space evolution and score-based solution space search. Solution space evolution entails using LLMs to evolve a solution space comprising multiple candidate solutions, rather than an individual solution. This approach aims to enhance the efficiency of LLM calls. Score-based solution space search leverages the scoring feedback from the evaluation function to guide the exploration of the solution space. Such scoring feedback is a crucial element that current LLMs often lack when independently tackling optimization problems.

\input{pic/spec_prompt}

\textbf{Task modeling.} An optimization problem is typically defined as finding an optimal solution \(x^* \in S\) for a given objective function \(f: S \rightarrow \mathbb{R}\), such that \(f(x^*) = \max_{x \in S} f(x)\). We use a task specification from FunSearch~\cite{romera2024mathematical} to outline the problem-solving process. 
Fig.~\hyperref[pic_spec_prompt]{\ref*{pic_spec_prompt}(a)} provides an example for the cap set problem, which involves constructing the largest possible subset in \(\mathbb{F}_3^n\) such that no three elements are collinear. The specification mainly includes three functions:
\begin{itemize}
\item Evaluate function, the objective function \(f\), which assesses whether the input \(x\) is a valid cap set and returns its size (i.e., score).
\item Solve function, which aims to identify an optimal solution \(x\) within the search space \(S\). We employ a greedy search, guided by a priority function, to approach this task. Although leveraging a greedy search with a priority function does not always guarantee the discovery of the optimal solution for all optimization problems, such problems can still be modeled in this manner. Therefore, designing an effective priority function is particularly important for solving optimization problems using this approach. 
\item Priority function, defined as $s_i = p(c_i, \sigma, \kappa)$, where $s_i$ denotes the priority score, $c_i$ represents the candidate choice, $\sigma$ signifies the current state (or partial solution), and $\kappa$ pertains to the global context. For reasons of computational efficiency and to avoid prohibitive time complexity, the input parameters $\sigma$ and $\kappa$ are sometimes omitted. In such scenarios, the score $s_i$ is calculated based solely on the intrinsic properties of the candidate $c_i$ itself. This priority function will be generated through our solution space evolution. We find that a good initial reference priority function can enhance final convergence. This will be discussed in detail in Sec.~\ref{initial_program}.
\end{itemize}

\textbf{Prompt building.} To instruct LLMs to generate priority functions, we first design the prompt. An example for the cap set problem is shown in Fig.~\hyperref[pic_spec_prompt]{\ref*{pic_spec_prompt}(b)}. The prompt consists of four main parts:
\begin{itemize}
\item Problem description, which briefly describes the problem being addressed. The problem definition may be included as needed.
\item Instructions, which include two key directives:
a) The evolution instruction, which instructs the model to develop an improved implementation based on the provided references. We do not precisely define “improvement”; instead, we hint that it “might lead to breakthroughs,” allowing the model more freedom to explore without requiring expert knowledge.
b) The tunable marker instruction, which instructs the model to identify tunable components.
\item Task description, which defines the specific requirements for the LLM’s generation.
\item Reference implementations, which include \(K_{\text{ref}}\) promising implementations sampled from a program database based on their scores. They provide a foundation for the model’s evolution. The sampling process will be introduced later.
\end{itemize}
While encouraging LLMs to generate more detailed descriptions can be useful~\cite{wei2022chain, kojima2022large}, it often results in progressively longer outputs and reduces control over code length in large-scale iterations. Therefore, we use the prompt “Output Python code only, without any comments.” to prohibit this.
We find that adding an evolutionary direction hint to the task description greatly helps final convergence. For instance, in the cap set and Shannon capacity problems, we use the prompt “The score is computed based on the relationships among el[i], el[-i], el[(i - k) \% n], and el[(i + k) \% n].” to guide the model toward symmetrical and neighborhood relationships. This will be discussed in detail in Sec.~\ref{initial_program}.

\textbf{LLM call.} LLMs act as the engine of evolution, inheriting discovered effective features by understanding reference implementations and mutating to produce new features by following evolutionary instructions. \(X\)-evolve primarily utilizes the Qwen2.5 72B Instruct~\cite{qwen2025qwen25technicalreport} and Llama 4 Maverick~\cite{meta2025llama} models via OpenRouter. Their pricing is comparable to Gemini 2.5 Flash~\cite{gemini2025gemini}, at approximately one-tenth the cost of Gemini 2.5 Pro~\cite{gemini2025gemini}. We find that as long as an LLM is well-trained and possesses sufficient instruction-following capabilities, its capacity has a limited impact on the final results. This observation aligns with findings from two other papers~\cite{romera2024mathematical, jordan2025generative}. The limited impact of LLM capacity is because for open, complex problems, the key factor driving evolution is not the model's prior knowledge, but its understanding of reference implementations and instructions. However, we also observe that some overly small models exhibit poor instruction-following. For instance, even when explicitly prohibiting descriptions and comments, their responses may still contain such content. Similarly, when instructed to use the tunable marker for tunable sections, they may fail to execute this directive effectively. This poor instruction adherence in such models significantly impacts experimental results. A more detailed discussion regarding LLM capacity is presented in Sec.~\ref{model_capacity}.

\textbf{\(X\)-search.}
Fig.~\hyperref[pic_introduction]{\ref*{pic_introduction}(b)} shows a simplified tunable program, and Program \hyperlink{pic_tunable_program}{1} in Appendix presents a complete program generated by an LLM for constructing the largest symmetric admissible set we found in \(\mathcal{A}(27,19)\). We use a program analysis tool to extract tunable structures, each representing an independent decision space. Their Cartesian product forms the solution space—for example, the one derived from the program in Program \hyperlink{pic_tunable_program}{1} contains about \(10^{15}\) possible solutions. Given the large solution space, we design a score-based space search algorithm. The main steps include:
\begin{itemize}
    \item Initial random sampling: iterate through all decision spaces, randomly sample a decision sequence, and replace each decision in the original tunable structure, converting the tunable program into a standard evaluable program. Repeat this process to generate a batch of evaluable programs.
    \item Batch evaluation: use the evaluate function from the specification to score each program in the batch. Assign these scores to each decision in the sequence. If a decision appears in multiple sequences, use the highest received score.
    \item Probabilistic sampling: in subsequent rounds, use the softmax function for probabilistic sampling, favoring higher-scoring paths. If a decision has never been sampled, assign it the maximum score of its decision space to encourage exploration.
    \item Termination condition: repeat the sampling and evaluation process until scores fail to improve for \(K_{\text{stall}}\) consecutive rounds.
\end{itemize}
After the space search, the tunable program is compacted by retaining only the decisions present in the top-\(K\) evaluable programs, removing all others. If a tunable has only one remaining option, remove the tunable construct and keep only that single choice. Finally, the compacted tunable program is added to the program database, and its score is recorded as the maximum score among the evaluated programs.

\textbf{Program database.}
Compacted tunable programs are added to the program database, with invalid and timeout-exceeding programs excluded. To sample reference programs from this database, a clustering-based sampling algorithm is used. The main steps are:
\begin{itemize}
    \item Cluster partitioning - Divide the database into \(K_{\text{cluster}}\) clusters: Programs with the top-1 score (possibly multiple programs) form a separate cluster. The remaining programs are grouped into the remaining \(K_{\text{cluster}}-1\) clusters based on their scores using the \(K\)-means algorithm~\cite{macqueen1967some}.
    \item Cluster sampling - Sample \(K_{\text{ref}}\) clusters from these \(K_{\text{cluster}}\) clusters. The probability \(p_i\) of selecting the \(i\)-th cluster is given by: \(p_i = p_0 \cdot e^{-\lambda \cdot i}\), where \(p_0\) is a predefined base probability (e.g., 0.5). The parameter \(\lambda\) is determined using binary search to satisfy \(\sum_{i=1}^k p_i = 1\).
    \item Program sampling - Randomly select one program from each of the chosen \(K_{\text{ref}}\) clusters as reference implementations for the next round of LLM evolution.
\end{itemize}
During the search process, the score growth curve typically rises quickly at first, then gradually slows down. In later stages, scores often stabilize within a certain range, leading to “score accumulation” (i.e., many programs clustering around a specific score range). When using a probability calculation function like softmax, this accumulation increases the likelihood of selecting programs from that range over time, as it dominates the probability distribution. This reinforces the accumulation and can trap the search in a local optimum. Score accumulation occurs because generated program scores tend to be close to one of their reference program scores. Clustering methods can help address this by grouping similar scores and sampling by cluster rather than individual scores, reducing oversampling in dense regions and avoiding local optima.
The top-1 score is placed in a separate cluster because the goal is to find the optimal solution, making it important to exploit the current best more effectively. Using the cluster index for probability calculation, rather than score statistics within clusters~\cite{romera2024mathematical}, offers better control. For instance, with \(p_0\) set to 0.5, the top three clusters have fixed sampling probabilities of 0.5, 0.25, and 0.13, totaling 0.85. In contrast, using score statistics can lead to scenarios where a sudden high-scoring mutation has a sampling probability over 0.99. Combined with score accumulation, this can cause the search process to get stuck. 
We sample programs randomly from a cluster rather than by program length~\cite{romera2024mathematical}, as we find that current LLMs tend to be verbose, making indirect length control through selecting short reference implementations ineffective. Instead, we remove the length constraint, and while this leads to slightly longer outputs, they remain within an acceptable range.

\textbf{Adaptive halving.}
To prevent getting stuck in local optima, we simultaneously initialize \(K_{\text{search}}\) search processes. After every \(K_{\text{reset}}\) model calls, the bottom half of the processes with the lowest scores is terminated (randomly selected in case of ties) and restarted. We prefer restarting rather than state copying from other processes~\cite{romera2024mathematical} for two main reasons: 1) Our method converges quickly, so restarting incurs only a modest additional cost compared to copying. 2) Copying can lead to data from a single process gradually contaminating all others over  \(\lceil \log_2 K_{\text{search}} \rceil\) copies, potentially reducing search diversity. If the source process lacks sufficient potential to evolve toward the optimal solution, this could negatively impact overall search quality. Ref.~\cite{jordan2025generative}, which says that 8 small-scale runs yield better results than a single large-scale run, makes an observation similar to ours. The ablation study is presented in Sec.~\ref{adaptive_halving}.

%% file: pic/overview_simple.tex
\begin{figure}[t]%
    \centering
    \includegraphics[scale=0.4]{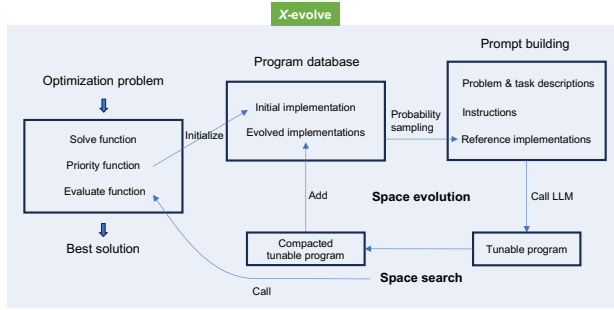} 

    \caption{\(X\)-evolve high-level overview.}%
    \label{pic_overview_simple}%

\end{figure}

%% file: pic/spec_prompt.tex
\begin{figure}[t]%
    \centering
    \includegraphics[scale=0.4]{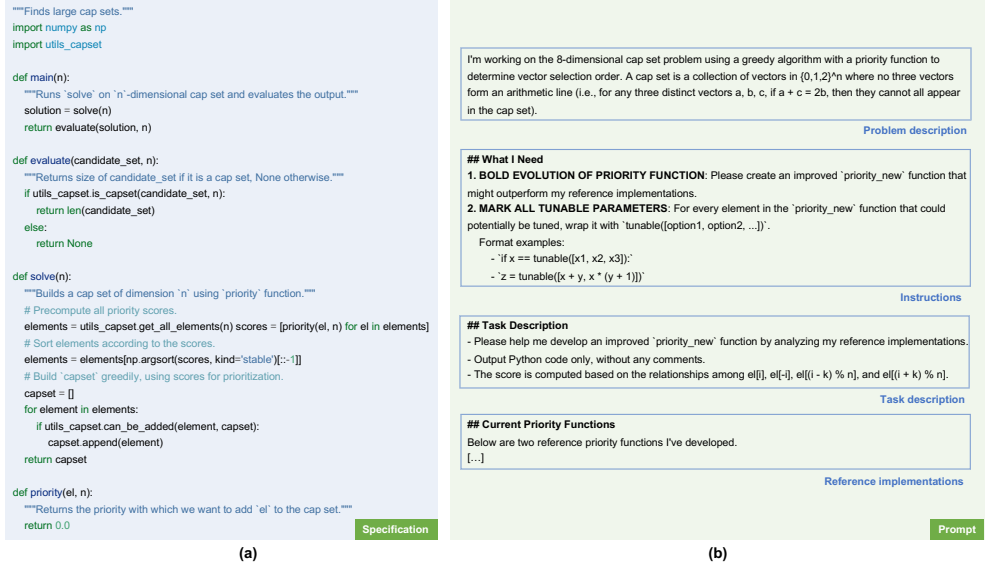} 

    \caption{Examples of \textbf{(a)} task specification and \textbf{(b)} prompt for the cap set problem. [...] denotes omitted content for brevity.}%
    \label{pic_spec_prompt}%

\end{figure}

%% file: 3_results.tex
\section{Results}\label{sec2}

\(X\)-evolve is well-suited for optimization problems that provide rich scoring feedback. We validate its effectiveness on three distinct challenges: the cap set problem, the Shannon capacity of cycle graphs, and the online bin packing problem. These challenging problems originate from the fields of combinatorics, graph theory, and operations research, respectively.

\input{3.1_cap_set}

\input{3.2_shannon}
\input{3.3_bin_packing}

%% file: 3.1_cap_set.tex
\subsection{Cap sets}

A cap set~\cite{grochow2019new, tao2006additive} in a finite \(d\)-dimensional vector space over a field \(\mathbb{F}_q\), typically \(\mathbb{F}_3\), is defined as a collection of points such that no three distinct points sum to zero (or, equivalently, form an affine line when \(\mathbb{F}_q = \mathbb{F}_3\)). 
In \(\mathbb{F}_3^2\), the “2” denotes the dimension (number of coordinates), and coordinates are taken from the finite field \(\mathbb{F}_3 = \{0, 1, 2\}\). For example, \((0, 0) + (1, 1) + (2, 2) = (3, 3)\), and \((3, 3) \pmod{3} = (0, 0)\), so these three points are collinear. Likewise, \((0, 0), (0, 1), (0, 2)\) and \((0, 1), (1, 2), (2, 0)\) are also collinear. The set \(\{(0, 0), (0, 1), (1, 0), (1, 1)\}\) is a maximal cap set in \(\mathbb{F}_3^2\), containing four points and no three collinear points.
This structure has attracted much research interest. This attention arises largely from the deceptively simple nature of its definition, which contrasts sharply with the profound difficulty of determining the maximum size of a cap set in \(\mathbb{F}_3^d\). 
Terence Tao has described the cap set problem as “perhaps my favourite open question”~\cite{terence2009open} and “a model problem for all these other questions in Ramsey theory”~\cite{erica2016simple}. In this work, we focus on two main aspects of the cap set problem: 1) determining the capacity \(C = \sup c_n^{1/n}\), a fundamental question in additive combinatorics~\cite{tao2006additive}; 2) finding the size \(c_n\) of the largest cap set in dimension \(n\).

\textbf{Lower bound.} Ref.~\cite{ellenberg2017large} advances the upper bound on the cap set constant \(C\) to \(C \leq 2.756\) using the polynomial method. Recent improvements on the lower bound stem from algorithmic innovations. Notably, FunSearch~\cite{romera2024mathematical} leverages an LLM+EA approach to raise the lower bound from 2.2180 to 2.2202—the largest improvement in 20 years. In this work, we push the bound further to 2.2203.

\input{table/admissible_set_results}

Following prior work~\cite{calderbank1994maximal, edel2004extensions, tyrrell2022new, romera2024mathematical}, our approach uses constant-weight admissible sets—admissible sets for short—as a key combinatorial tool in cap set construction. Specifically, we construct a larger cap set and derive a lower bound through the following process. 1) 
Construct a symmetric admissible set \( \mathcal{A}(n, w) \), consisting of vectors in \( \{0, 1, 2\}^n \). The full admissible set, denoted \( \mathcal{I}(n, w) \), has size \( \binom{n}{w} \). The structure of these vectors is invariant under independent cyclic permutations of coordinates within four disjoint groups of coordinate triples. This symmetry is discovered by FunSearch through analysis of its solution-generating program.
2) Expand the symmetric admissible set into an admissible set of size \(s\). 3) Using the admissible set, we construct a cap set \(A \subseteq \mathbb{F}_3^{6 \cdot m \cdot n}\). The size of \(A\) is given by \(s \cdot b_0^{n - w} \cdot b_1^w\), where \(b_0 = a_0 \cdot m \cdot a_1^{m - 1}\), \(b_1 = a_1^m\), and \(m\) (e.g., 4, 5, 7) is taken from a recursively defined admissible set \(I(m, m - 1)\), with \((a_0, a_1) = (12, 112)\) derived from two 6-dimensional cap sets. This approach reduces the search space from cap sets to admissible sets, and further to symmetric admissible set construction. The resulting smaller and more structured search space enables exploration of higher dimensions than was previously possible.
We define the score of a symmetric admissible set as its size. To greedily construct a symmetric admissible set with a higher score (i.e., larger size), we use \(X\)-evolve to evolve a priority function: \(\{0, 1, 2\}^n \rightarrow \mathbb{R}\). The results are presented in Table~\ref{table_admissible_set_results}.

We identify a full-size admissible set $\mathcal{I}(15,10)$ of size 3{,}003 with a lower bound $C = 2.2194$, and a set $\mathcal{A}(24,17)$ of size 237{,}984 with $C = 2.2202$—both matching the best known results from FunSearch.  
We also construct a partial admissible set $\mathcal{A}(21,15)$ of size 43{,}650 with $C = 2.2200$, exceeding the previous best known size of 43{,}596.  
In higher dimensions unexplored by FunSearch, we discover a partial admissible set $\mathcal{A}(27,19)$ of size 1{,}270{,}863, establishing a new lower bound for the cap set problem: $C = 2.2203$.
The $X$-evolve approach offers clear advantages: it first uses score-based feedback to guide the search, and then leverages the LLM to efficiently evolve the solution space. This process enables rapid convergence and significantly reduces search costs. As a result, we typically achieve high-quality solutions with fewer than 20{,}000 LLM calls---sometimes as few as a few hundred. This efficiency enables the exploration of higher dimensions, which was previously computationally prohibitive.
Notably, we observe diminishing improvements in the lower bound as the dimension increases: an increase of 0.0006 from dimension 15 to 21, 0.0002 from 21 to 24, and only 0.0001 from 24 to 27. These diminishing returns suggest that achieving further improvements in higher dimensions may become increasingly challenging.

\textbf{Largest cap sets.} Constructing large cap sets is more challenging than deriving lower bounds, primarily due to the vast search space. For instance, brute-forcing an 8-dimensional cap set of size 512 involves exploring \(\binom{3^8}{512} \approx 10^{779}\) possibilities. For perspective, while the estimated number of particles in the universe is about \(10^{80}\), the game of Go has a search space of approximately \(250^{150} \approx 10^{360}\)~\cite{silver2016mastering}.

\input{table/cap_set_results}

We address this problem using a greedy search strategy guided by a heuristic priority function, optimized via \(X\)-evolve. As shown in Table~\ref{table_cap_set_results}, our results match the state-of-the-art (FunSearch) in dimensions 3 through 8. In 8 dimensions, FunSearch finds a cap set of size 512 only 4 times out of 140 runs, whereas our method succeeds 3 times out of 5 (see Appendix \hyperlink{pic_capset_all}{A}), demonstrating superior search efficiency.

%% file: table/admissible_set_results.tex
\begin{table}[t!]
\centering
\begin{tabular}{ccc}
\toprule
\textbf{Lower bound on $C$} & \textbf{Admissible set ingredient} & \textbf{Source} \\
\midrule
2.2101 & $\mathcal{I}(90, 89)$ & Ref.~\cite{calderbank1994maximal} \\
2.2173 & $\mathcal{I}(10, 5)$  & Ref.~\cite{edel2004extensions} \\
2.2180 & $\mathcal{I}(11, 7)$  & Ref.~\cite{tyrrell2022new} \\
\midrule
2.2194 & $\mathcal{I}(15, 10)$ & FunSearch, $X$-evolve \\
2.2200 & $\mathcal{A}(21, 15)$  & $X$-evolve \\
2.2202 & $\mathcal{A}(24, 17)$ & FunSearch, $X$-evolve \\
2.2203 & $\mathcal{A}(27, 19)$ & $X$-evolve \\
\bottomrule
\end{tabular}
\caption{Summary of lower bounds on the cap set capacity $C$.}
\label{table_admissible_set_results}
\end{table}

%% file: table/cap_set_results.tex
\begin{table}[t!]
\centering
\begin{tabular}{ccccccc}
\toprule
$n$ & 3 & 4 & 5 & 6 & 7 & 8 \\
\midrule
Best known & 9 & 20 & 45 & 112 & 236 & 496 \\
FunSearch  & 9 & 20 & 45 & 112 & 236 & 512 \\
$X$-evolve  & 9 & 20 & 45 & 112 & 236 & 512 \\
\bottomrule
\end{tabular}
\caption{Sizes of the largest cap sets in \(\mathbb{F}_3^n\) for different dimensions $n$. The best known row excludes results from FunSearch.}
\label{table_cap_set_results}
\end{table}

%% file: 3.2_shannon.tex
\subsection{Shannon capacity of cycle graphs}

The Shannon capacity of a graph is an important invariant in information theory~\cite{shannon1956zero, korner1998zero}. It characterizes the asymptotic rate of error-free information transmission over multiple uses of a noisy communication channel~\cite{alon1998shannon, zuiddam2019asymptotic}.  
Let $G = (V_g, E_g)$ and $H = (V_h, E_h)$ be two graphs. The strong product $G \boxtimes H$ is the graph with vertex set $V_g \times V_h$, where two distinct vertices $(v_{g_1}, v_{h_1})$ and $(v_{g_2}, v_{h_2})$ are adjacent if ($v_{g_1} = v_{g_2}$ or $v_{g_1}v_{g_2} \in E_g$) and ($v_{h_1} = v_{h_2}$ or $v_{h_1}v_{h_2} \in E_h$). An independent set is a subset of vertices where no two are adjacent.
The strong product $G^{\boxtimes n}$ is defined recursively as the graph obtained by taking the strong product of $G$ with itself $n$ times. The independence number $\alpha(G^{\boxtimes n})$ denotes the size of the largest independent set in $G^{\boxtimes n}$. The Shannon capacity of a graph $G$ is defined as the exponential growth rate of the independence number, formally given by $\Theta(G) = \sup_n \sqrt[n]{\alpha(G^{\boxtimes n})}$.

In the study of Shannon capacity, cycle graphs $\mathcal{C}_{m}$ (with $m$ vertices connected in a closed chain) constitute one of the most extensively investigated classes of graphs, owing to their natural appearance in modeling communication scenarios with symmetric, cyclic confusability structures. A classic example is the 5-cycle $\mathcal{C}_{5}$, which models a communication channel with five equidistant signal values, where each symbol may be confused with its neighbors due to noise. Despite their simple and regular structure, determining the Shannon capacities of cycle graphs, particularly for odd cycles, has proven to be a challenging problem~\cite{bohman2003limit, godsil1995problems}. To date, the only odd cycle for which the Shannon capacity is exactly known is $\mathcal{C}_{5}$, for which $\Theta(\mathcal{C}_{5}) = \sqrt{5}$ was determined using the Lovász theta function~\cite{lovasz1979shannon}. For other odd cycles, the exact Shannon capacity remains unknown; only bounds are available, and refining these estimates continues to be an active area of research.

We apply the proposed \(X\){-evolve} framework to search for a priority function that assigns a priority score to each vertex in the cycle graph. In the subsequent greedy algorithm, vertices with higher scores are prioritized when constructing the independent set. This approach aims to produce larger independent sets in the strong product graph $\mathcal{C}_{m}^{\boxtimes n}$, thereby improving the lower bounds on the Shannon capacities of these graphs. Our results are summarized as follows.

\vspace{0.5em}
\noindent
\textbf{Recover best results}
\begin{itemize}
    \item For the 9-vertex cycle graph, we identify priority functions that yield maximum independent sets of sizes 81, 324, 1,458, 6,561, and 26,244 when applied to the strong product graphs $\mathcal{C}_{9}^{\boxtimes n}$ for $n=3,\ldots,7$, respectively—matching the best known results to date~\cite{mathew2017new}.
    \item For the 11-vertex cycle graph, we find a program that recovers the known lower bound of Shannon capacity in $\mathcal{C}_{11}^{\boxtimes 3}$, achieving an independent set of size 148~\cite{baumert1971combinatorial}. We also identify a priority function that produces a maximum independent set of size 754 in the strong product $\mathcal{C}_{11}^{\boxtimes 4}$, matching the best known result~\cite{romera2024mathematical}. The program, shown in Appendix \hyperlink{pic_cyclic_program_1}{A}, is composed of modular arithmetic, linear combinations, and simple nonlinear terms, exhibiting strong structural regularity and interpretability.
    \item For the 13-vertex cycle graph, we recover the known lower bound in $\mathcal{C}_{13}^{\boxtimes 3}$, achieving an independent set of size 247. Furthermore, we identify a priority function recovering the best known result of $\alpha(\mathcal{C}_{13}^{\boxtimes 5}) \ge 9,633$~\cite{baumert1971combinatorial}.
\end{itemize}

\noindent
\textbf{Improve lower bound}
\begin{itemize}
    \item For the 15-vertex cycle graph, we identify a priority function that constructs a larger independent set in the strong product $\mathcal{C}_{15}^{\boxtimes 5}$, improving the best known lower bound to $\alpha(\mathcal{C}_{15}^{\boxtimes 5}) \ge 19,946$. The previous best known lower bound was 19,894~\cite{de2024asymptotic}. The program, shown in Appendix \hyperlink{pic_cyclic_program}{A}, is concise and interpretable, potentially facilitating further research on cycle graphs.
\end{itemize}

%% file: 3.3_bin_packing.tex
\subsection{Bin packing}

The bin packing problem~\cite{coffman1984approximation} is a classic NP-hard problem in computer science. Its goal is to pack a set of items with varying sizes into the fewest number of fixed-capacity bins. Bin packing is divided into offline and online versions, depending on whether all item information is known beforehand. We focus on the online version, where items arrive sequentially, and each must be placed immediately without knowledge of future items. This real-time constraint increases uncertainty and makes optimization more difficult.

To address this problem, both heuristic algorithms~\cite{burke2006evolving, burke2007automatic, romera2024mathematical} and learning-based methods~\cite{wang2025bin, zhang2021attend2pack, hu2017solving, zhao2021online}—including deep learning (DL) and reinforcement learning (RL)—have been explored. Regardless of the approach, each defines a decision policy \(\pi\), whose effectiveness depends on its ability to generalize to unseen item sequences. While DL/RL models have shown promise, heuristics still offer key advantages:
\begin{itemize}
    \item Interpretability: Heuristics use clear, rule-based logic that is easy to understand and debug. This allows integration of expert knowledge and better handling of edge cases, unlike the “black-box” nature of DL/RL models.
    \item Low cost and easy deployment: They require little computation and can run efficiently on standard CPUs, avoiding the need for expensive hardware like GPUs or TPUs.
    \item Robustness: DL is a fundamentally numerical approximation technique. DL models’ rich parameterization offers strong fitting capability but also leads to a key drawback—overfitting: they often excel on training data but struggle in real-world applications~\cite{zhang2016understanding, recht2019imagenet}. In contrast, heuristics may not always achieve globally optimal solutions, but their simplicity and rule-based nature generally lead to more stable and reliable performance.
\end{itemize}
Two widely used heuristic algorithms are first fit~\cite{burke2006evolving} and best fit~\cite{burke2007automatic}. First fit places each item in the first bin with enough space. If no such bin exists, it opens a new bin. Best fit chooses the bin that leaves the least remaining space after placement and opens a new bin if needed.

\input{table/bin_packing_results}

The way we use \(X\)-evolve to solve the bin packing problem differs from its application to the cap set and Shannon capacity problems. For bin packing, the objective is to evolve a policy \(\pi\) with strong generalization ability, while the latter problems focus on finding an optimal solution without emphasizing generalization. Here, the policy \(\pi\) is actually a heuristic function, similar to the priority function used in the cap set and Shannon capacity problems. To evolve the policy \(\pi\), we follow a standard deep learning pipeline: evolving the policy on a training set, selecting the best-performing version on a validation set, and finally evaluating the evolved policy on the test set. 

We begin by evaluating the effectiveness of \(X\)-evolve on the well-known bin packing benchmarks from the OR-Library~\cite{beasley1990or}. The benchmark suite consists of four datasets (OR1 to OR4), each containing 20 instances with item counts of 100, 250, 500, and 1,000, respectively. To ensure the training, validation, and test sets are independently and identically distributed (i.i.d.), we generate a training set using the same sampling strategy as in the OR-Library construction process, matching the scale of OR1 (20 instances, each with 100 items) and a validation set matching the scale of OR2 (20 instances, each with 250 items). See Appendix~\ref{binpacking_details} for additional details.

The score for the bin packing problem is defined as the proportion of excess bins used by the current policy, relative to the $L_2$ lower bound~\cite{martello1990lower} of the optimal offline bin packing solution. Table~\ref{bin_packing_result} shows the results: \(X\)-evolve outperforms both first fit and best fit on all datasets, achieving performance comparable to FunSearch~\cite{romera2024mathematical} while requiring only about 100 LLM calls. The results indicate that the heuristic policy learned by \(X\)-evolve generalizes well: even when trained on smaller instances such as OR1, it still performs effectively on larger test instances (i.e., OR2 to OR4). The heuristic succeeds because it blends a smooth “best-fit” preference with a set of sharply tuned boosts and penalties. The rule first ignores bins that can’t hold the item, then steadily amplifies the appeal of bins the nearer they are to full. A sharp nonlinear reward and decisive “almost-full” bonuses drive each piece into the tightest existing space before any new bin is opened, yielding reliably high packing efficiency on the OR-Library benchmarks. For the heuristic function and further analysis, please refer to Appendix \hyperlink{pic_binpacking_OR}{A}.

The Weibull(\(k\), \(\lambda\)) distribution’s flexible shape \(k\) and scale \(\lambda\) parameters allow it to model a wide range of real-world patterns, particularly varying failure rates. This makes it a valuable tool for analyzing reliability, material strength, and wind speed. In addition to the bin packing benchmarks from the OR-Library, we also evaluate the effectiveness of \(X\)-evolve on instances sampled from a Weibull distribution (see Appendix~\ref{binpacking_details} for details), as the Weibull distribution closely fits the data in many real-world scheduling problems~\cite{castineiras2012weibull, angelopoulos2023online}. Table~\ref{bin_packing_result} presents the results: \(X\)-evolve performs excellently on the test sets, significantly outperforming both first fit and best fit, while also demonstrating strong generalization ability. The heuristic works because it swiftly packs small, common Weibull items into near-full bins while sparing a few empty bins for the rare large ones—minimizing gaps and new-bin openings at the same time. For the heuristic function and further analysis, please refer to Appendix \hyperlink{pic_binpacking_Weibull}{A}.

%% file: table/bin_packing_results.tex
\begin{table}[t!]
\centering
\begin{tabular}{cccccccc}
\toprule
& \textbf{OR1} & \textbf{OR2} & \textbf{OR3} & \textbf{OR4} & \textbf{Weibull} & \textbf{Weibull} & \textbf{Weibull} \\
&&&&& \textbf{5k} & \textbf{10k} & \textbf{100k} \\
\midrule
First fit & 6.42\% & 6.45\% & 5.74\% & 5.23\% & 4.17\% & 4.17\% & 3.95\% \\
\midrule
Best fit & 5.81\% & 6.06\% & 5.37\% & 4.94\% & 3.88\% & 3.82\% & 3.75\% \\
\midrule
FunSearch & \textbf{5.30\%} & \textbf{4.19\%} & 3.11\% & 2.47\% & - & - & - \\
\midrule
\textbf{\(X\)-evolve} & 5.50\% &  4.43\% &  \textbf{2.98\%} &  \textbf{2.45\%} &  \textbf{0.67\%} &  \textbf{0.38\%} &  \textbf{0.12\%} \\
\bottomrule
\end{tabular}
\caption{The fraction of excess bins in online bin packing. A smaller fraction indicates better performance. \(X\)-evolve outperforms first fit and best fit across two benchmarks and is comparable to FunSearch. The priority function discovered by FunSearch for the Weibull distribution is not open-source, so the corresponding data is unavailable.
}
\label{bin_packing_result}
\end{table}

%% file: 4_ablations.tex
\section{Ablations}\label{ablations}

\input{4.1_scale_of_model}
\input{4.2_tunable}
\input{4.3_adaptive_halving}

\input{4.4_initial_prompt}

\subsection{Summary}\label{ablation_summary}

We do not directly compare the model call overhead with FunSearch for two main reasons. First, FunSearch is not fully open-sourced and cannot be run out of the box. Second, its primary model, Codey—an LLM built on the PaLM2 family—is no longer available. Nevertheless, a rough comparison is still possible. The FunSearch paper reports using around \(10^6\) samples in total, and its supplementary material shows that in smaller-dimension experiments (e.g., on the symmetric operable set $\mathcal{I}(15,10)$), the model was called approximately 2.5 million times. In contrast, our method typically requires only around \(10^4\) model calls across the entire paper. As shown in Sec.~\ref{model_capacity}, it remains effective even when using an 8B-parameter model. This suggests that our approach reduces LLM calls by roughly two orders of magnitude compared to prior state-of-the-art methods.

\(X\)-evolve explores a space rather than a single point in each iteration, significantly reducing LLM call costs. Take the cap set problem as an example: each model call roughly involves 2,000 input tokens and 1,000 output tokens. 
Based on Gemini 2.5 Flash pricing, 20{,}000 calls would cost approximately
\((2000 \times 0.15 / 10^6 + 1000 \times 0.6 / 10^6) \times 20000 = 18 \text{ USD}\),
where 0.15 and 0.6 refer to the cost per million input and output tokens, respectively, from OpenRouter. Qwen 2.5 72B and LLaMA 4 Maverick offer similar pricing. As a result, we bring LLM usage within reach for typical researchers, helping advance LLM+EA approaches for open-ended, complex problems. However, this also highlights a new challenge. As discussed in Sec.~\ref{x_search}, while model call frequency has dropped significantly, evaluation calls remain high and may become a new bottleneck—an issue that calls for further research.

%% file: 4.1_scale_of_model.tex
\subsection{Impact of model capacity on search efficiency}\label{model_capacity}

We begin by examining how model capacity affects the search efficiency of \(X\)-evolve. Specifically, we employ six LLMs of varying sizes to search for the Shannon capacity of the graph \(\mathcal{C}_{13}^{\boxtimes 5}\). These models include a mixture-of-experts (MoE) model and dense models ranging in size from 72B to 8B. Detailed specifications are provided in Table~\ref{tab:abl1}.

\input{table/c_13_5result}

The ablation results are shown in Table~\ref{tab:abl1}. We summarize the following key findings:
\begin{itemize}
\item For all LLMs except Deepseek R1 Distill Llama 8B, search performance remains largely consistent across models, suggesting that it is not strongly tied to model capacity. This is likely because search effectiveness is driven by prompt (reference implementations and instructions) understanding—rather than prior knowledge. This observation aligns with findings from two related studies~\cite{romera2024mathematical, jordan2025generative}.
\item For the Deepseek R1 Distill Llama 8B, we find that it lacks reliable instruction-following capabilities. This is evidenced by the following: even when explicitly instructed to avoid descriptions or comments, the model may still generate such content; and when asked to use tunable markers to indicate tunable sections, it often fails to comply effectively. This highlights the necessity of using a well-trained, instruction-following LLM in the \(X\)-evolve pipeline.
\item Both Qwen Coder models (32B and 7B) perform consistently well, reinforcing that models fine-tuned on code are better suited to this task than general-purpose LLMs. Notably, FunSearch also leverages code-specialized models—Codey and StarCoder (15B).
\item Each LLM has its own inherent biases, which may cause the search process to converge prematurely to local optima. To address this, we experiment with hybrid model calls (e.g., LLaMA 4 Maverick + Qwen2.5 72B), aiming to introduce cross-model perturbations. However, this strategy yields no significant improvement in performance.
\end{itemize}

%% file: table/c_13_5result.tex
\begin{table}[t!]
    \centering
    \begin{tabular}{|c|c|c|c|c|}
        \hline
        \textbf{Model} & \textbf{Run 1} & \textbf{Run 2} & \textbf{Run 3} & \textbf{Mean} \\ \hline 

        Llama 4 Maverick~\cite{meta2025llama} & 1287 & 2141 & 1037 & 1488 \\ \hline
        Qwen2.5 72B Instruct~\cite{qwen2025qwen25technicalreport} & 736 & 1211 & 4199 & 2049 \\ \hline
        \makecell[c]{Llama 4 Maverick\\+ Qwen2.5 72B Instruct} & 2494 & 4055 & 722 & 2424 \\ \hline
        Qwen2.5 Coder 32B Instruct~\cite{hui2024qwen2} & 1132 & 503 & 2507 & 1381 \\ \hline
        Llama 3 8B Instruct~\cite{grattafiori2024llama} & 2077 & 456 & 3680 & 2071 \\ \hline
        Deepseek R1 Distill Llama 8B~\cite{guo2025deepseek} & 10000 (8892) & 10000 (9126) & 10000 (9126) & 10000 \\ \hline
        Qwen2.5 Coder 7B Instruct~\cite{hui2024qwen2} & 147 & 1061 & 1181 & 796 \\ \hline
    \end{tabular}
    \caption{
    Ablation results on model capacity for the Shannon capacity problem in $\mathcal{C}_{13}^{\boxtimes 5}$. LLMs are ordered from top to bottom by decreasing parameter count. Llama 4 Maverick is a MoE model with 400B total parameters and 17B active per call. The row “Llama 4 Maverick + Qwen2.5 72B Instruct” indicates that each LLM call randomly chooses one of the two models. Experiments are repeated with three random seeds (42, 2025, 1013), each given a 10,000 LLM call budget. Values indicate the number of calls needed to reach the best-known $\alpha(\mathcal{C}_{13}^{\boxtimes 5})$ (i.e., an independent set of size 9633). If the best-known value is not reached within the budget, the final achieved size is shown in parentheses. The “Mean” column reports the average number of calls across the three runs.
    }
    \label{tab:abl1}
\end{table}

%% file: 4.2_tunable.tex
\subsection{Impact of the tunable marker}\label{x_search}

In the \(X\)-evolve framework, a key feature is leveraging the tunable marker instruction to prompt LLMs to generate tunable programs. From each of these tunable programs, we parse a solution space, $X$, and then search this space for the best possible solution. Our search process iteratively samples and evaluates a batch of evaluable programs from $X$. This cycle repeats until the evaluation score fails to improve for $K_{\text{stall}}$ consecutive rounds. Therefore, a single call to the LLM results in $\text{batch\_size} \times k$ evaluations, where $\text{batch\_size}$ is the number of evaluable programs in a batch and $k$ (with \(k > K_{\text{stall}}\)) is the total number of evaluation rounds.

To test the efficiency of this method, we run three experiments on the Shannon capacity of the graph $\mathcal{C}_{13}^{\boxtimes 5}$, comparing the number of LLM calls and total evaluations required. The experiments are: 1) Without the tunable marker, where each LLM call generates only a single evaluable program; 2) With the tunable marker, using a $\text{batch\_size}$ of 8: total evaluations = \(8 \times k\); and 3) With the tunable marker, using a $\text{batch\_size}$ of 64: total evaluations = \(64 \times k\). The LLM call budgets are set to 50,000 for the first experiment and 10,000 for the other two.

\input{pic/x_search}

Fig.~\ref{fig:x_search1} illustrates how the score (i.e., maximum independent set size) increases with LLM call times. The experiments using the tunable marker show a much faster score increase compared to the baseline (without the tunable marker). They converge to high scores (\(\geq\)9,126) within fewer than 11 calls, whereas the baseline requires more than 2,854 calls. This demonstrates that tunable program generation significantly reduces the cost of LLM calls. Moreover, a larger \(\text{batch\_size}\) (64) accelerates convergence compared to a smaller one (8), indicating that larger evaluation batches better explore the solution space. Fig.~\ref{fig:x_search2} shows score progression with respect to evaluation times. In the baseline, the LLM call limit of 50,000 bounds the number of evaluations. In contrast, with the tunable marker, a single tunable program can yield a large number of evaluable programs, resulting in evaluations reaching the millions. Overall, the tunable marker slightly improves evaluation times but does not lead to a significant reduction. This reveals a potential issue: when LLM calls are no longer the system bottleneck, evaluation itself might become the new bottleneck—an aspect that requires further research. Additionally, with tunable markers, \(\text{batch\_size}\)=8 outperforms \(\text{batch\_size}\)=64, implying that smaller batches may be more cost-effective, similar to small-batch advantages in neural network training. When evaluations are under 1,000, a batch size of 64 even underperforms the baseline, indicating that excessively large batches may introduce an evaluation burden. Taken together, from both Fig.\ref{fig:x_search1} and Fig.\ref{fig:x_search2}, while \(\text{batch\_size}\)=64 cannot lower evaluation costs, it substantially reduces LLM call costs. The trade-off of larger batch sizes is generally advantageous, as a single LLM call is far more expensive than an evaluation, and in our experience, the evaluation overhead when \(\text{batch\_size}\)=64 remains within an acceptable range in some cases.

%% file: pic/x_search.tex
\begin{figure}
    \centering
    \subfigure[]{
    \label{fig:x_search1}
    \includegraphics[width=0.4\linewidth]{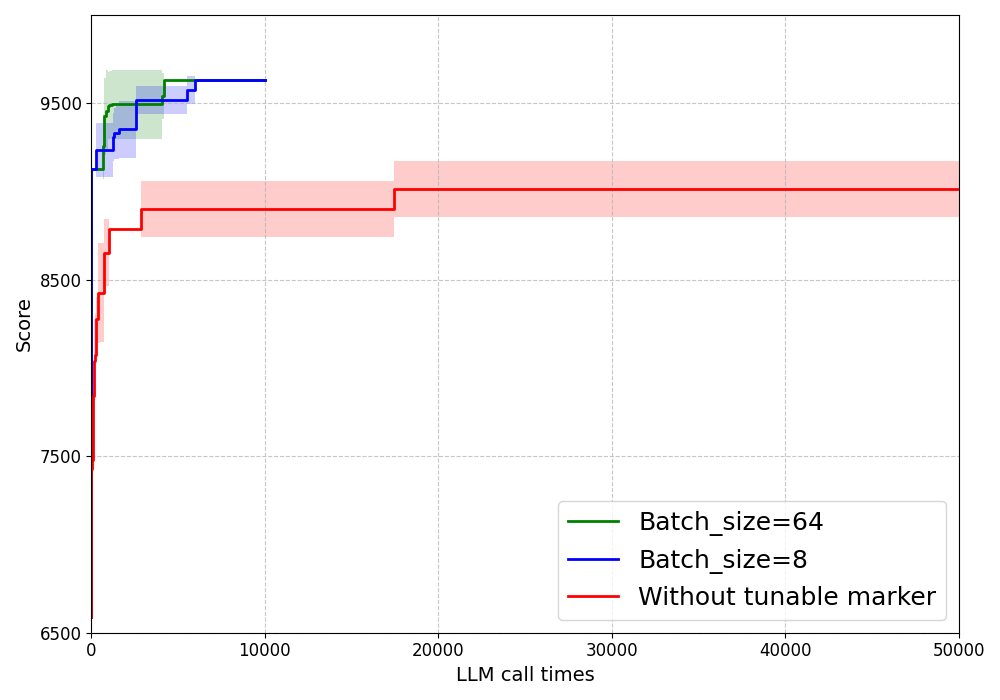}
    }
    \subfigure[]{
    \label{fig:x_search2}
    \includegraphics[width=0.4\linewidth]{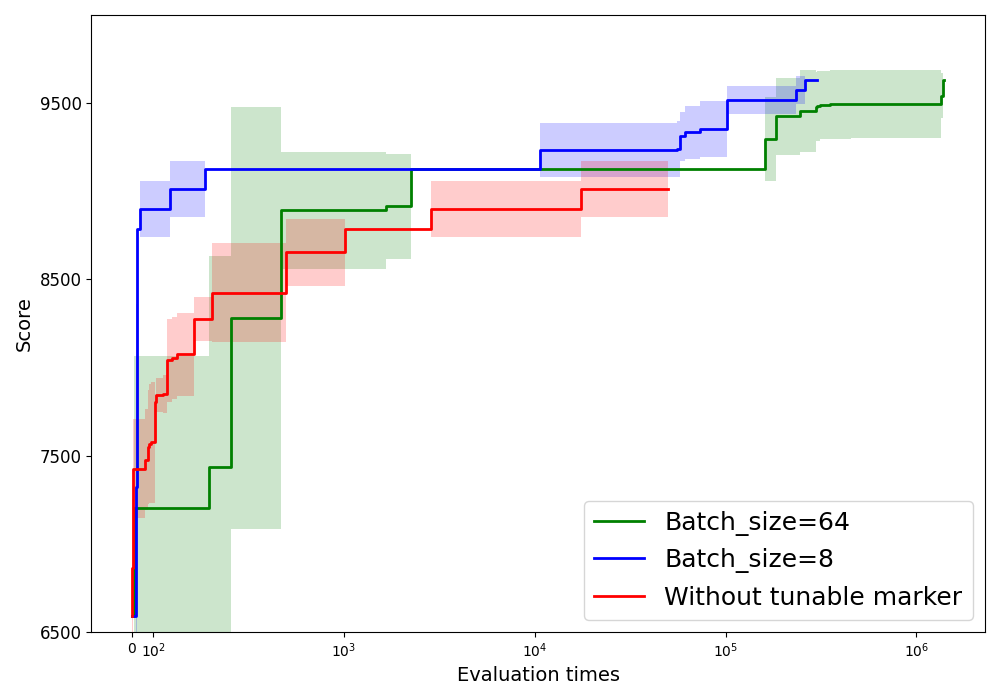}
    }
    \caption{Ablation study on the effect of the tunable marker on search efficiency. All experiments are conducted with the Qwen2.5 72B Instruct model. When the experimental setting specifies \(\text{batch\_size}\)=64, it indicates that the tunable marker is activated, and 64 programs are sampled and evaluated per iteration. Similarly, a configuration with \(\text{batch\_size}\)=8 implies that the tunable marker is active, but only 8 programs are sampled and evaluated per iteration. In contrast, the experiment setting "without tunable marker" represents cases where the tunable marker is not used. \textbf{(a)} The score's growth curve relative to the number of LLM calls. The \(x\)-axis corresponds to the number of calls made to the LLM, while the \(y\)-axis represents the size of the largest independent set achieved. Results are averaged over three independent runs. The experiments utilize LLM call budgets of 10,000, 10,000, and 50,000, respectively. \textbf{(b)} The score's growth curve relative to the number of evaluations. The \(x\)-axis, which is presented on a logarithmic scale, indicates the number of evaluations, and the \(y\)-axis shows the size of the largest independent set. In the setting without the tunable marker, the maximum number of model calls allowed is 50,000, so the plot is capped at 50,000 on the \(x\)-axis. For the \(\text{batch\_size}\)=8 setting, the best average result occurs at 261,501 evaluations, while for \(\text{batch\_size}\)=64, the best average result is achieved at 1,372,816 evaluations.
    }
    \label{fig:x_search}
\end{figure}

%% file: 4.3_adaptive_halving.tex
\subsection{Adaptive halving}\label{adaptive_halving}

The adaptive halving strategy is an important component of the \(X\)-evolve framework. To avoid getting stuck in local optima, we launch \(K_\text{search}\) parallel search processes. After every \(K_\text{reset}\) LLM calls (counting all processes in total), the bottom half of the processes—those with the lowest scores—are terminated and restarted. As shown in the previous section, using tunable marker and space search, the search process quickly converges to high-scoring solutions. Instead of copying states from better-performing processes—which can reduce search diversity—we simply restart under-performing ones. Although this introduces some overhead, it is minimal and acceptable. To evaluate the effectiveness of adaptive halving in escaping local optima, we conduct experiments on the 8-dimensional cap set problem, with the LLM call budget = 20,000 and \(K_\text{search}\) = 4. We compare two settings: one with adaptive halving (\(K_\text{reset}\) = 3200) and one without. Each setting is repeated for five runs, and the results are shown in Table~\ref{tab:adaptive_halving}. With adaptive halving, three out of five runs reach the current best-known score of 512. In contrast, all five runs without it are trapped in local optima and fail to achieve the optimum. In more detail, during Run 4 with adaptive halving, the best solution is generated by search process 1, which has already been reset twice. In Run 2, process 0 is reset four times before producing a score of 464, but is terminated when the call budget is exhausted—suggesting it might have reached 512 if allowed to continue.

\input{table/adaptive_halving}

The effectiveness of adaptive halving can be attributed to the inherent randomness in LLM outputs, which leads each search process to explore different regions of the search space. When progress stalls, adaptive halving resets the process, allowing fresh exploration. Meanwhile, better-performing processes continue evolving until they either converge to an optimal solution or stagnate and are reset.

%% file: table/adaptive_halving.tex
\begin{table}[t!]
    \centering
    \begin{tabular}{|c|c|c|c|c|c|}
        \hline
         Settings & Run1 & Run2 & Run3 & Run4 & Run5\\ \hline
         Adaptive halving with \(K_\text{reset}\) = 3200 & 512 & 464 & 512 & 512 & 449\\ \hline
         Without adaptive halving & 450 & 496 & 464 & 464 & 450\\
         \hline
    \end{tabular}
    \caption{Adaptive halving results for the 8-dimensional cap set problem, with the LLM budget limited to 20,000.
    }
    \label{tab:adaptive_halving}
\end{table}

%% file: 4.4_initial_prompt.tex
\subsection{Search direction guidance}\label{initial_program}

We find that effective guidance of the search direction is crucial to the performance of \(X\)-evolve. This guidance takes two main forms. First, natural language cues can be added to the prompt’s task description—for example, a \textit{scoring heuristic}: “The score is computed based on the relationships among el[i], el[-i], el[(i - k) \% n], and el[(i + k) \% n].” Such hints encourage the model to focus on symmetry and neighborhood structure. Second, a well-crafted initial priority function can be imported into the specification. While low-dimensional priority functions may not scale directly to higher dimensions, they often capture valuable problem-solving heuristics and can thus be surprisingly effective. Importantly, such functions are generally easier to obtain. For example, in the cap set problem, the maximum size in the first six dimensions is known and can be explicitly constructed. Additionally, low-dimensional search spaces are typically small enough to allow even brute-force enumeration.

We run three sets of experiments on both the Shannon capacity problem and the symmetric admissible set problem, each with a model call budget of 10,000. The configurations include: no guidance, natural language guidance via the \textit{scoring heuristic}, and guidance using an initial priority function defined in the specification.

\input{pic/init}

As shown in Fig.~\ref{fig:init}, for the Shannon capacity problem, using an initial priority function significantly accelerates the search, whereas natural language guidance has a negative impact. In contrast, for the symmetric admissible set problem, both types of guidance improve performance, with natural language cues performing slightly better (i.e., reaching the best score more quickly). These results suggest that selecting an appropriate guidance strategy for different types of problems can have a substantial impact on search efficiency.

In this paper, we adopt the \(\mathcal{C}_9\) priority function from FunSearch—a concise and elegant design—for all Shannon capacity experiments, except for \(\mathcal{C}_9\) itself, which uses FunSearch’s \(\mathcal{C}_7\) priority function. For the cap set and symmetric admissible set problems, we incorporate natural language guidance via the \textit{scoring heuristic} descriptions.

%% file: pic/init.tex
\begin{figure}
    \centering
    \subfigure[Shannon capacity in $\mathcal{C}_{13}^{\boxtimes 5}$]{
    \label{fig:init1}
    \includegraphics[width=0.4\linewidth]{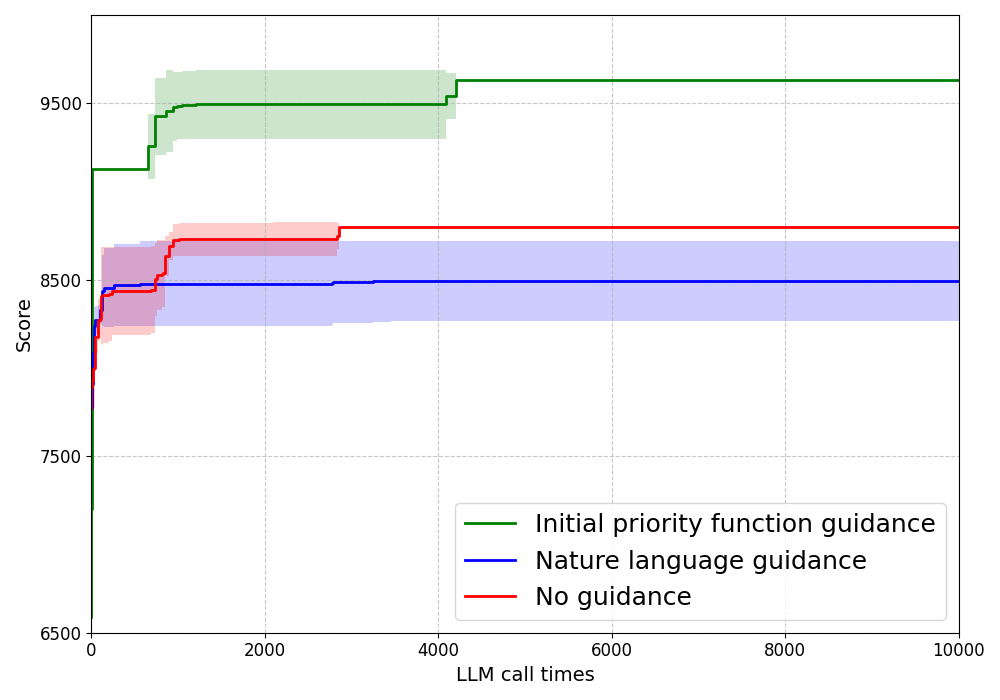}
    }
    \subfigure[Symmetry admissible set in $\mathcal{A}(21,15)$]{
    \label{fig:init2}
    \includegraphics[width=0.4\linewidth]{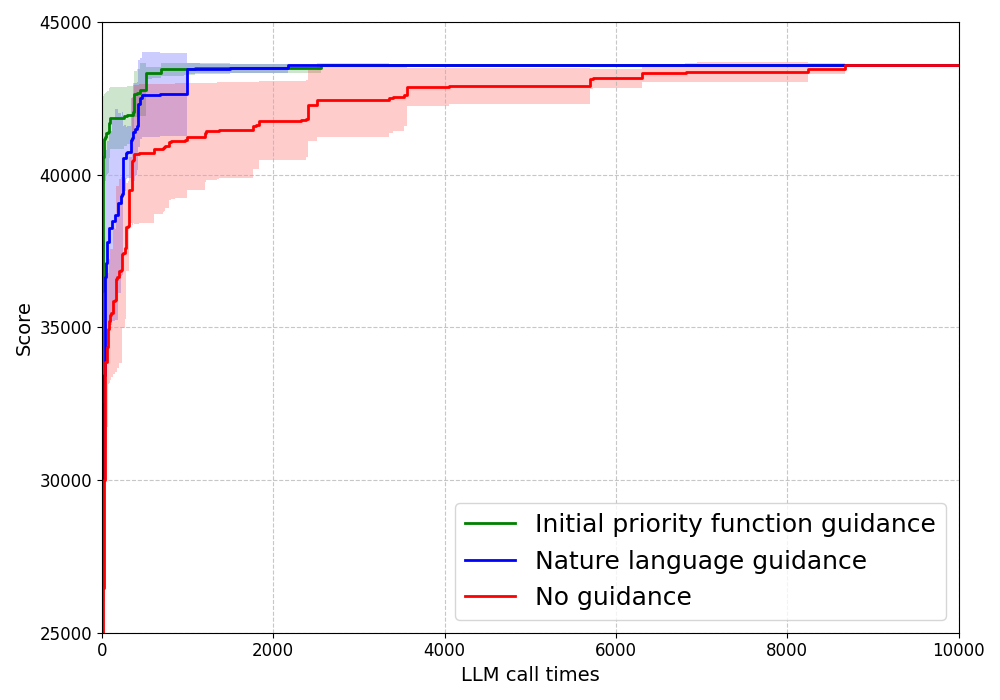}
    }
    \caption{
Ablation results of the search direction guidance. All experiments are conducted using the Qwen2.5 72B Instruct model, with a fixed call budget of 10,000 for each experiment. 
\textbf{(a)} Results on the Shannon capacity of the cycle \(\mathcal{C}_{13}^{\boxtimes 5}\). The \(x\)-axis represents the number of LLM calls, while the \(y\)-axis indicates the size of the largest independent set in the graph \(\mathcal{C}_{13}^{\boxtimes 5}\). The natural language guidance specifies: “The score is computed based on the relationships among el[i], el[-i], el[(i - k) \% n], and el[(i + k) \% n].” The initial priority function guidance utilizes the \(\mathcal{C}_9\) priority function provided by FunSearch. 
\textbf{(b)} Results on the symmetry admissible set \(\mathcal{A}(21, 15)\). The \(x\)-axis represents the number of LLM calls, and the \(y\)-axis denotes the size of the largest admissible set in \(\mathcal{A}(21, 15)\). The natural language guidance used here remains consistent with that of the Shannon Capacity experiments, and the initial priority function uses the priority function of \(\mathcal{I}(15, 10)\), provided by FunSearch.
}
    \label{fig:init}
\end{figure}

%% file: 5_relate.tex
\section{Related work}

Genetic programming (GP) is an evolutionary computation technique inspired by Darwinian natural selection. It evolves computer programs by exploring a space defined by functions and terminals to solve specific tasks or approximate target functions. Since its inception~\cite{koza1994genetic, banzhaf1998genetic}, GP has been widely applied in symbolic regression~\cite{koza1994genetic2, vladislavleva2008order, schmidt2009distilling, ma2022evolving}, complex function discovery~\cite{langdon2013foundations}, automated modeling~\cite{banzhaf1998genetic, o2010open}, optimization algorithm design~\cite{chen2023symbolic}, and control strategy synthesis~\cite{koza1999genetic, jakobovic2006dynamic, zhang2021genetic}. Despite these successes, GP faces notable challenges~\cite{o2010open}, particularly in defining the search space and designing effective operators. Successful mutation relies on well-defined transformation rules, which often require problem-specific structures and expert insight. Although several approaches~\cite{koza1994genetic, salustowicz1997probabilistic} have sought to mitigate this, developing general, robust mutation operators remains a significant hurdle.

The rapid advancement of Large Language Models (LLMs) has greatly facilitated their integration with evolutionary algorithms (EAs). A notable example is DeepMind’s FunSearch~\cite{romera2024mathematical}, which combines an LLM with an EA framework to tackle challenges in both pure mathematics and applied computer science. Since its introduction, FunSearch has been successfully applied to a range of downstream tasks, including Bayesian optimization~\cite{aglietti2024funbo}, cognitive model discovery~\cite{castro2025discovering}, graph distance computation~\cite{verma2025grail}, and combinatorial programming~\cite{velivckovic2024amplifying}. More recently, Ref.~\cite{jordan2025generative} presents an updated version of FunSearch designed to support mathematicians who may lack expertise in machine learning or access to high-performance computing resources.
AlphaEvolve~\cite{alpha2025alexander} is a substantial enhancement of FunSearch by DeepMind. Rather than generating entire programs, it produces diff blocks—akin to git-style conflict markers—to suggest localized code edits. This approach is particularly well-suited for large codebases, as it reduces the output burden on the LLM and avoids the risk of unintended changes to unrelated code.

In recent years, LLMs have demonstrated significant promise in the field of software engineering, with applications spanning from routine development tasks to highly specialized domains. They have been successfully applied to standalone programming challenges~\cite{austin2021program, quan2025codeelo} and to common workflows such as resolving GitHub issues~\cite{jimenez2023swe}. More recently, there has been growing interest in their capability to generate GPU kernel code in domain-specific languages like CUDA~\cite{ouyang2025kernelbench}—a particularly impactful direction given both the performance-critical nature of such code and the limited pool of expert developers in this area. Some efforts focus on training stronger code models~\cite{hui2024qwen2, roziere2023code, zhu2024deepseek}, while others enhance output quality by leveraging additional compute during inference. LLMs have also been explored for optimizing key GPU operations, such as attention modules~\cite{chen2025deepseek} and user-defined functions~\cite{lange2025ai}. Beyond these use cases, recent work has investigated their potential in discovering novel evolutionary algorithms~\cite{lange2024large}, as well as in assisting with the training of LLMs themselves~\cite{lehman2023evolution}. Taken together, these diverse lines of research underscore the versatility of LLMs and their expanding role in shaping the future of software engineering.

%% file: 6_discussion.tex
\section{Discussion}

\textbf{New discoveries.} \(X\)-evolve evolves genuinely new structures, rather than merely retrieving results from existing libraries. We construct symmetric admissible sets of sizes 43,650 and 1,270,683 in \(\mathcal{A}(21, 15)\) and \(\mathcal{A}(27, 19)\), respectively; a maximum independent set of size 19,946 in the 15-vertex cycle graph; and three distinct 8-dimensional cap sets of size 512. Several of these results surpass the best known records and represent novel discoveries beyond the scope of any LLM training data.

\textbf{RL.} \(X\)-evolve shares similarities with reinforcement learning (RL). The programs generated by \(X\)-evolve can be seen as policies in RL. \(X\)-evolve produces priority scores for candidate choices, while RL typically outputs a probability distribution over actions. Both are then used to guide decision-making. Their difference lies in their optimization methods—\(X\)-evolve relies on evolutionary algorithms, whereas RL usually employs methods like gradient descent. Importantly, with the rapid progress of large language models (LLMs), their well-trained natural language understanding makes the evolutionary process more accessible. By “accessible,” we mean that the evolutionary process can now be described in natural language rather than explicitly coded. These similarities raise several questions worth exploring:
\begin{itemize}
\item \textit{Can evolutionary algorithms be used to train RL models?} Yes. As shown in Ref.~\cite{salimans2017evolution}, although traditionally viewed as less effective for RL, well-designed evolution strategies can rival state-of-the-art RL algorithms and offer superior scalability.
\item \textit{Can LLMs’ world knowledge support RL training?} This is an active and promising area of research, with LLMs being explored for various roles such as reward design~\cite{ma2023eureka, kwon2023reward}, high-level planning~\cite{du2023guiding}, grounding actions in affordances~\cite{ahn2022can, brohan2023rt}, and guiding exploration~\cite{wang2023describe}.
\item \textit{Can gradient descent be used to optimize \(X\)-evolve for a given problem?} Directly optimizing LLMs with gradient descent is costly and not recommended. However, since \(X\)-evolve explores the search space through sampling, introducing a differentiable decision model might enable more efficient exploration via gradient-based optimization.
\end{itemize}

\textbf{AlphaEvolve.} AlphaEvolve~\cite{alpha2025alexander} is a substantial enhancement of FunSearch by DeepMind. Rather than generating entire programs, it produces diff blocks—akin to git-style conflict markers—to suggest localized code edits. This approach is particularly well-suited for large codebases, as it reduces the output burden on the LLM and avoids the risk of unintended changes to unrelated code. As for its relation to our \(X\)-evolve, there are two main points to consider: 1) AlphaEvolve offers limited immediate benefits for the specific problems we are addressing (e.g., cap set, Shannon capacity of cycle graphs), as our code is relatively short and does not necessitate fine-grained diffs. As their paper notes: “Where the code being evolved is very short… AlphaEvolve can output the entire code block directly.” 2) AlphaEvolve and \(X\)-evolve are orthogonal in nature. Looking ahead, as our framework \(X\)-evolve scales to handle more complex or expansive codebases, adopting a diff-based strategy with tunable markers to designate optimization regions could enhance convergence and significantly reduce the effective code search space.

%% file: 7_acknowledgements.tex
\newpage
\section*{Data availability}

The experiments in this study utilize the publicly available OR-Library bin packing benchmarks~\cite{beasley1990or}. The programs produced by \(X\)-evolve are presented in Appendix~\ref{discovered_program}.

\section*{Code availability}
The source code of this study is available at GitHub (\href{https://github.com/zhaiyi000/x-evolve}{https://github.com/zhaiyi000/x-evolve}).

\section*{Acknowledgements}
This work was partially supported by the National Natural Science Foundation of China (No. 624B2138, No. 62332016, and No. 62272434). This research was also supported by the advanced computing resources provided by the Supercomputing Center of the USTC.

\section*{Author contributions}
Y.Z. designed the research. Y.Z., Z.Q.W., and R.H.L. conducted the experiments. All authors analyzed the results. Y.Z., Z.Q.W., R.H.L., W.Y.Z., and Y.Y.Z. wrote the manuscript. All authors reviewed the manuscript.

\section*{Competing interests}

The authors plan to file a patent related to this work, with the University of Science and Technology of China as their primary affiliation.

%% file: a_discovered_program.tex
\newpage
\section{Discovered programs}\label{discovered_program}
This section presents several programs discovered by \(X\)-evolve.
\\
\\
\input{program/tunable_program}
\newpage
\input{program/admissible_set_1}
\newpage
\input{program/admissible_set}
\newpage
\newpage
\hypertarget{pic_capset_all}{}
\noindent
Program \hyperlink{pic_capset_1}{4}, Program \hyperlink{pic_capset_2}{5}, and Program \hyperlink{pic_capset_3}{6} are functions discovered by \(X\)-evolve, each constructing a 512-cap in \(n=8\) dimensions, which matches the previously known best construction obtained by Funsearch. Those programs exhibit significant differences in length and structure, reflecting the strengths of \(X\)-evolve in terms of diversity and exploration.
\\
\\
\input{program/512_1}
\newpage
\input{program/512_2}
\newpage
\input{program/512_3}
\newpage
\input{program/cyc11_4}
\\
\\
\input{program/cyclic_graphs}
\newpage
\input{program/binpacking_or}
\\
\\
\input{program/binpacking_weibull}

%% file: program/tunable_program.tex
\hypertarget{pic_tunable_program}{}
\definecolor{codegreen}{rgb}{0,0.6,0}
\definecolor{codeblue}{rgb}{0,0,1}
\definecolor{codered}{rgb}{0.6,0,0}
\lstdefinestyle{custompython}{
    language=Python,
    basicstyle=\ttfamily\fontsize{6pt}{8pt}\selectfont,
    keywordstyle=\color{codegreen},
    identifierstyle=\color{codeblue},
    commentstyle=\color{codered}\itshape,
    stringstyle=\color{codered}\itshape,
    showstringspaces=false,
    numbers=none,
    frame=none,
    breaklines=true,
    tabsize=4,
    aboveskip=0em,    
    belowskip=0em,
    basewidth=0.5em,
    lineskip=-1pt        
}

\noindent
\textbf{Program 1:} The tunable program generated by an LLM for constructing the largest symmetric admissible set we found in \(\mathcal{A}(27,19)\). In the final stages of the \(X\)-evolve framework search, program structural tuning ceases, and the emphasis shifts primarily to parameter tuning. The solution space, as parsed from the tunable program, contains \(3^{31} \approx 6.1 \times 10^{15}\) solutions.

\noindent\hrulefill
\begin{lstlisting}[style=custompython]
def priority_new(el: tuple[int, ...], n: int, w: int) -> float:
    k_values = [1, 2, 2, 3, 4, 4, 2, 5, 5, 5]
    score = 0.0
    count_1 = sum(tunable([0.5, 0.6, 0.7]) if x == 1 else 0 for x in el)
    count_2 = sum(tunable([1.0, 1.1, 1.2]) if x == 2 else 0 for x in el)
    count_0 = sum(tunable([0.5, 0.6, 0.7]) if x == 0 else 0 for x in el)
    for i in range(n):
        if el[i] == 1:
            score += tunable([0.0, 0.1, -0.1]) * (el[(n - i) % n] == 2)
            score += tunable([3.7, 4.0, 4.3]) * (el[(i - k_values[0]) % n] == 0)
            score += tunable([4.0, 4.5, 5.0]) * (el[(i + k_values[0]) % n] == 2)
            score += tunable([0.5, 0.6, 0.7]) * (el[(i + k_values[1]) % n] == el[i])
            score += tunable([-1.5, -1.0, -2.0]) * (el[(i - k_values[1]) % n] == el[(i + k_values[1]) % n])
            score += tunable([0.6, 0.7, 0.8]) * ((el[(i + k_values[0]) % n] + el[(i - k_values[0]) % n]) == 3)
            score += tunable([-0.7, -0.8, -0.6]) * (el[(i + k_values[2]) % n] == el[(i - k_values[2]) % n])
            score += tunable([0.8, 0.9, 1.0]) * (el[(i + 2 * k_values[0]) % n] == 2)
            score += tunable([0.9, 1.0, 1.1]) * (el[(i - 2 * k_values[0]) % n] == 0)
            score += tunable([1.0, 1.1, 1.2]) * (el[(i + k_values[0] + k_values[1]) % n] == el[i])
            score += tunable([0.2, 0.3, 0.4]) * (el[(i - k_values[0] - k_values[1]) % n] == \
            el[(i + k_values[0] + k_values[1]) % n])
            score += tunable([0.3, 0.4, 0.5]) * (el[(i + k_values[3]) % n] == 1)
            score += tunable([0.4, 0.5, 0.6]) * (el[(i - k_values[3]) % n] == 1)
            score += tunable([0.1, 0.2, 0.3]) * (el[(i + k_values[4]) % n] == 2)
            score += tunable([0.5, 0.6, 0.7]) * (el[(i - k_values[4]) % n] == 2)
            score += tunable([0.1, 0.2, 0.3]) * (el[(i + k_values[5]) % n] == 0)
            score += tunable([-0.2, -0.1, -0.3]) * (el[(i - k_values[5]) % n] == 0)
            score += tunable([0.5, 0.6, 0.7]) * (el[(i + k_values[6]) % n] == 1)
            score += tunable([0.0, 0.1, -0.1]) * (el[(i + k_values[7]) % n] == 0)
            score += tunable([0.7, 0.8, 0.9]) * (el[(i + k_values[8]) % n] == 2)
            score += tunable([0.0, 0.1, -0.1]) * (el[(i + k_values[9]) % n] == 0)
    normalization = w ** tunable([0.6, 0.7, 0.8])
    score /= normalization
    score += tunable([0.4, 0.5, 0.6]) * (count_1 / w)
    score += tunable([0.6, 0.7, 0.8]) * (count_2 / (n - w))
    score += tunable([0.1, 0.15, 0.2]) * (count_0 / (n - w))
    score *= tunable([1.8, 2.0, 2.2])
    score += tunable([0.2, 0.3, 0.4]) * ((count_1 + count_2) / n)
    score += tunable([1.5, 1.6, 1.7]) * (sum(1 for x in el if x == 0) / n)
    return score
\end{lstlisting}
\noindent\hrulefill

%% file: program/admissible_set_1.tex
\hypertarget{pic_admiss_program_1}{}
\definecolor{codegreen}{rgb}{0,0.6,0}
\definecolor{codeblue}{rgb}{0,0,1}
\definecolor{codered}{rgb}{0.6,0,0}
\lstdefinestyle{custompython}{
    language=Python,
    basicstyle=\ttfamily\fontsize{6pt}{8pt}\selectfont,
    keywordstyle=\color{codegreen},
    identifierstyle=\color{codeblue},
    commentstyle=\color{codered}\itshape,
    stringstyle=\color{codered}\itshape,
    showstringspaces=false,
    numbers=none,
    frame=none,
    breaklines=true,
    tabsize=4,
    aboveskip=0em,      
    belowskip=0em,
    basewidth=0.5em,
    lineskip=-1pt        
}

\noindent
\textbf{Program 2:} The function discovered by \(X\)-evolve that builds a size 43,650 admissible set in \(\mathcal{A}(21,15)\). This is larger than the best known admissible set in \(\mathcal{A}(21,15)\) obtained by Funsearch, which has size 43,596.

\noindent\hrulefill
\begin{lstlisting}[style=custompython]
def priority(el: tuple[int, ...], n: int, w: int) -> float:
    score = 0.0
    for i, val in enumerate(el):
        if val == 1:
            score += 2.5
        elif val == 2:
            score += 5.0
        else:
            score += -0.8
        if i % 3 == 0:
            score += 2.0
        if i % 5 == 0:
            score += -1.2
        if i % 7 == 0:
            score += 1.4
    if sum(el) == w:
        score *= 1.6
    else:
        score *= 1.8
    score += 0.6 * (w - sum(1 for x in el if x != 0))
    score += 0.0 * sum(el[i] * el[(i + 1) % n] for i in range(n))
    score -= 1.6 * sum(el[i] * el[(i + 2) % n] for i in range(n))
    score += 0.9 * sum(el[i] * el[(i + 3) % n] for i in range(n))
    score -= 0.6 * sum(el[i] * el[(i + 4) % n] for i in range(n))
    score += 0.9 * sum(el[i] * el[(i + 5) % n] for i in range(n))
    score -= 0.5 * sum(el[i] * el[(i + 6) % n] for i in range(n))
    score += 0.4 * sum(el[i] * el[(i + 7) % n] for i in range(n))
    score -= 0.8 * sum(el[i] * el[(i + 8) % n] for i in range(n))
    score += 0.8 * sum(el[i] * el[(i + 9) % n] for i in range(n))  
    score -= 0.9 * sum(el[i] * el[(i + 10) % n] for i in range(n))
    score += 0.7 * sum(el[i] * el[(i + 11) % n] for i in range(n))
    score -= 0.3 * sum(el[i] * el[(i + 12) % n] for i in range(n))   
    score += 0.4 * sum(el[i] * el[(i + 13) % n] for i in range(n))
    score -= 0.7 * sum(el[i] * el[(i + 14) % n] for i in range(n))
    score += 0.9 * sum(el[i] * el[(i + 15) % n] for i in range(n))
    score -= 1.1 * sum(el[i] * el[(i + 16) % n] for i in range(n))
    score += 0.5 * sum(el[i] * el[(i + 17) % n] for i in range(n))
    score -= 0.5 * sum(el[i] * el[(i + 18) % n] for i in range(n))
    score += 0.6 * sum(el[i] * el[(i + 19) % n] for i in range(n))
    score -= 0.9 * sum(el[i] * el[(i + 20) % n] for i in range(n))
    if sum(el) == 14:
        score += 3.5
    else:
        score -= 4.0
    return score
\end{lstlisting}
\noindent\hrulefill

%% file: program/admissible_set.tex
\hypertarget{pic_admiss_program}{}
\definecolor{codegreen}{rgb}{0,0.6,0}
\definecolor{codeblue}{rgb}{0,0,1}
\definecolor{codered}{rgb}{0.6,0,0}
\lstdefinestyle{custompython}{
    language=Python,
    basicstyle=\ttfamily\fontsize{6pt}{8pt}\selectfont,
    keywordstyle=\color{codegreen},
    identifierstyle=\color{codeblue},
    commentstyle=\color{codered}\itshape,
    stringstyle=\color{codered}\itshape,
    showstringspaces=false,
    numbers=none,
    frame=none,
    breaklines=true,
    tabsize=4,
    aboveskip=0em,     
    belowskip=0em,
    basewidth=0.5em,
    lineskip=-1pt       
}

\noindent
\textbf{Program 3:} The function discovered by \(X\)-evolve that builds a size 1,270,863 admissible set in \(\mathcal{A}(27,19)\). This admissible set implies an improved lower bound of 2.2203 on the cap set capacity, which exceeds the best known lower bound of 2.2202 obtained by Funsearch.

\noindent\hrulefill
\begin{lstlisting}[style=custompython]
def priority(el: tuple[int, ...], n: int, w: int) -> float:
    k_values = [1, 2, 2, 3, 4, 4, 2, 5, 5, 5]
    score = 0.0
    count_1 = sum(0.7 if x == 1 else 0 for x in el)
    count_2 = sum(1.0 if x == 2 else 0 for x in el)
    count_0 = sum(0.5 if x == 0 else 0 for x in el)
    for i in range(n):
        if el[i] == 1:
            score += 0.1 * (el[(n - i) % n] == 2)
            score += 4.3 * (el[(i - k_values[0]) % n] == 0)
            score += 4.5 * (el[(i + k_values[0]) % n] == 2)
            score += 0.5 * (el[(i + k_values[1]) % n] == el[i])
            score += -1.0 * (el[(i - k_values[1]) % n] == el[(i + k_values[1]) % n])
            score += 0.6 * ((el[(i + k_values[0]) % n] + el[(i - k_values[0]) % n]) == 3)
            score += -0.8 * (el[(i + k_values[2]) % n] == el[(i - k_values[2]) % n])
            score += 1.0 * (el[(i + 2 * k_values[0]) % n] == 2)
            score += 1.0 * (el[(i - 2 * k_values[0]) % n] == 0)
            score += 1.0 * (el[(i + k_values[0] + k_values[1]) % n] == el[i])
            score += 0.2 * (el[(i - k_values[0] - k_values[1]) % n] == el[(i +
                           k_values[0] + k_values[1]) % n])
            score += 0.4 * (el[(i + k_values[3]) % n] == 1)
            score += 0.4 * (el[(i - k_values[3]) % n] == 1)
            score += 0.3 * (el[(i + k_values[4]) % n] == 2)
            score += 0.7 * (el[(i - k_values[4]) % n] == 2)
            score += 0.1 * (el[(i + k_values[5]) % n] == 0)
            score += -0.1 * (el[(i - k_values[5]) % n] == 0)
            score += 0.7 * (el[(i + k_values[6]) % n] == 1)
            score += 0.1 * (el[(i + k_values[7]) % n] == 0)
            score += 0.9 * (el[(i + k_values[8]) % n] == 2)
            score += 0.1 * (el[(i + k_values[9]) % n] == 0)
    normalization = w ** 0.6
    score /= normalization
    score += 0.4 * (count_1 / w)
    score += 0.8 * (count_2 / (n - w))
    score += 0.2 * (count_0 / (n - w))
    score *= 2.2
    score += 0.4 * ((count_1 + count_2) / n)
    score += 1.6 * (sum(1 for x in el if x == 0) / n)
    return score
\end{lstlisting}
\noindent\hrulefill

%% file: program/512_1.tex
\hypertarget{pic_capset_1}{}
\definecolor{codegreen}{rgb}{0,0.6,0}
\definecolor{codeblue}{rgb}{0,0,1}
\definecolor{codered}{rgb}{0.6,0,0}
\lstdefinestyle{custompython}{
    language=Python,
    basicstyle=\ttfamily\fontsize{6pt}{8pt}\selectfont, 
    keywordstyle=\color{codegreen},
    identifierstyle=\color{codeblue},
    commentstyle=\color{codered}\itshape,
    stringstyle=\color{codered}\itshape,
    showstringspaces=false,
    numbers=none,
    frame=none,
    breaklines=true,
    tabsize=4,
    aboveskip=0em,      
    belowskip=0em,
    basewidth=0.5em,
    lineskip=-1pt         
}

\noindent
\textbf{Program 4:} The function discovered by \(X\)-evolve that builds a 512-cap in \(n=8\) dimensions.

\noindent\hrulefill
\begin{lstlisting}[style=custompython]
def priority(el: tuple[int, ...], n: int) -> float:
    score = 2.5
    for i in range(n):
        k = 5
        weight_i = 1.0 if el[i] == 0 else -1.0
        weight_symmetry = 2.0 if el[i] == el[-i - 1] else -3.0
        weight_k_symmetry = 0.05 if el[(i - k) % n] == el[(i + k) % n] else 0.005
        score += weight_i
        score += weight_symmetry
        score += weight_k_symmetry
        aggregate = el[i] + el[-i - 1] + el[(i - k) % n] + el[(i + k) % n]
        score += 0.3 * aggregate
        product = el[i] * el[-i - 1] * el[(i - k) % n] * el[(i + k) % n]
        score += 0.01 * product
        condition_score = 0.15 * (
            (el[i] == 1) & 
            (el[(i - k) % n] == 1) & 
            (el[(i + k) % n] == 1)
        )
        score += condition_score
    parity_factor = 2.0 if sum(el) % 2 == 1 else 0.8
    score *= parity_factor
    score += 0.0
    score = score * 0.8
    return score
\end{lstlisting}
\noindent\hrulefill

%% file: program/512_2.tex
\hypertarget{pic_capset_2}{}
\definecolor{codegreen}{rgb}{0,0.6,0}
\definecolor{codeblue}{rgb}{0,0,1}
\definecolor{codered}{rgb}{0.6,0,0}
\lstdefinestyle{custompython}{
    language=Python,
    basicstyle=\ttfamily\fontsize{6pt}{8pt}\selectfont, 
    keywordstyle=\color{codegreen},
    identifierstyle=\color{codeblue},
    commentstyle=\color{codered}\itshape,
    stringstyle=\color{codered}\itshape,
    showstringspaces=false,
    numbers=none,
    frame=none,
    breaklines=true,
    tabsize=4,
    aboveskip=0em,     
    belowskip=0em,
    basewidth=0.5em,
    lineskip=-1pt         
}

\noindent
\textbf{Program 5:} The function discovered by \(X\)-evolve that builds a 512-cap in \(n=8\) dimensions.

\noindent\hrulefill
\begin{lstlisting}[style=custompython]
def priority(el: tuple[int, ...], n: int) -> float:
    score = 0.0
    k = 2
    weight_symmetry = -1.0
    weight_neighbor = 0.2
    antipodal_offset = n // 2
    for i in range(n):
        neighbor1 = el[(i + k) % n]
        neighbor2 = el[(i - k) % n]
        antipodal = el[(i + antipodal_offset) % n]
        symmetry_score = 0.3 if el[i] == antipodal else 0.0
        neighbor_score = 0.2 if neighbor1 == neighbor2 else -0.3
        score += weight_symmetry * symmetry_score + weight_neighbor * neighbor_score
        ops = [
            (el[i] == 1),
            (el[(i - k) % n] == 0),
            (el[(i + k) % n] == 0)
        ]
        condition1 = ops[0] & ops[1] & ops[2]
        score += 0.3 * condition1
        condition2 = (el[i] == antipodal) & (el[(i - k) % n] == el[(i + k) % n])
        score += -2.0 * condition2
        condition3 = (el[i] + el[(i + n // 3) % n] == 4) & (el[(i - k) % n] + el[(i + k) % n] == 2)
        score += 0.4 * condition3
        score += 0.6 * (el[i] == 1)
        score += -0.1 * abs(el[i] - antipodal)
        score += 0.01 * abs(el[i] - neighbor1)
        score += -0.01 * abs(el[i] - neighbor2)
        score += -0.01 * (el[i] == neighbor1)
        score += 0.0 * (el[i] == neighbor2)
        score += -0.02 * (el[i] == el[(i + k + 1) % n])
        score += 0.01 * abs(el[i] - el[(i + k + 2) % n])
    score += 0.4 * (sum(el) % 2)
    score *= 4.0
    return score
\end{lstlisting}
\noindent\hrulefill

%% file: program/512_3.tex
\hypertarget{pic_capset_3}{}
\definecolor{codegreen}{rgb}{0,0.6,0}
\definecolor{codeblue}{rgb}{0,0,1}
\definecolor{codered}{rgb}{0.6,0,0}
\lstdefinestyle{custompython}{
    language=Python,
    basicstyle=\ttfamily\fontsize{6pt}{8pt}\selectfont,
    keywordstyle=\color{codegreen},
    identifierstyle=\color{codeblue},
    commentstyle=\color{codered}\itshape,
    stringstyle=\color{codered}\itshape,
    showstringspaces=false,
    numbers=none,
    frame=none,
    breaklines=true,
    tabsize=4,
    aboveskip=0em,   
    belowskip=0em,
    basewidth=0.5em,
    lineskip=-1pt      
}

\noindent
\textbf{Program 6:} The function discovered by \(X\)-evolve that builds a 512-cap in \(n=8\) dimensions.

\noindent\hrulefill
\begin{lstlisting}[style=custompython]
def priority(el: tuple[int, ...], n: int) -> float:
    score = 0.0
    weights = [0.1, 0.001, 0.0005, 0.02, 0.001, 0.4, 0.02,
                     0.01, 0.02, 0.05, 2.0, 0.8, 0.01, 0.001, 0.005]
    ks = [4, 6]
    ops = ['!=', 'add']
    modes = ['cyclic']
    multipliers = [1.0, 1.0]
    mods = [2, 8]
    for i in range(n):
        if modes[0] == 'normal':
            x = el[i]
            y = el[-i - 1]
        elif modes[0] == 'cyclic':
            x = el[i]
            y = el[(i + 1) % n]
        else:
            x = el[i]
            y = el[i - 1]
        k1 = ks[0]
        k2 = ks[1]
        if modes[0] == 'normal':
            xk1 = el[(i - k1) % n]
            yk1 = el[(i + k1) % n]
            xk2 = el[(i - k2) % n]
            yk2 = el[(i + k2) % n]
        elif modes[0] == 'cyclic':
            xk1 = el[(i + k1) % n]
            yk1 = el[(i - k1) % n]
            xk2 = el[(i + k2) % n]
            yk2 = el[(i - k2) % n]
        else:
            xk1 = el[(i - k1) % n]
            yk1 = el[(i - k1 - 1) % n]
            xk2 = el[(i - k2) % n]
            yk2 = el[(i - k2 - 1) % n]
        if ops[0] == '!=':
            score += weights[0] * (y != 0)
        elif ops[0] == '==':
            score += weights[0] * (y == 0)
        elif ops[0] == '<':
            score += weights[0] * (y < 1)
        else:
            score += weights[0] * (y > 0)
        if ops[1] == 'add':
            score += weights[1] * (xk1 + yk1)
        elif ops[1] == 'sub':
            score += weights[1] * (xk1 - yk1)
        else:
            score += weights[1] * (xk1 * yk1)
        score -= weights[2] * abs(xk1 + yk1 - 0 * x)
        score += weights[3] * (xk2 == yk2)
        score -= weights[4] * abs(xk2 + yk2 - 2 * x)
        score += weights[5] * (x == 1)
    score *= multipliers[0]
    score += multipliers[1] * (sum(el) % mods[0])
    score -= weights[6] * abs(sum(el) - n) ** 3
    score += weights[7] * (sum(el) % mods[1])
    score += weights[8] * (sum([el[i] == el[-i - 1] for i in range(n)]) % mods[0])
    score += weights[9] * (sum([el[i] != el[-i - 1] for i in range(n)]) % mods[1])
    score += weights[10] * (sum([el[i] == el[(i + k1) % n] for i in range(n)]) % mods[0])
    score += weights[11] * (sum([el[i] != el[(i + k1) % n] for i in range(n)]) % mods[1])
    score += weights[12] * (sum([el[i] == el[(i + k2) % n] for i in range(n)]) % mods[0])
    score += weights[13] * (sum([el[i] != el[(i + k2) % n] for i in range(n)]) % mods[1])
    score += weights[14] * (sum([el[i] == (el[(i + k1) % n] + el[(i + k2) % n]) // 2 for i in range(n)]) % mods[0])
    return score

\end{lstlisting}
\noindent\hrulefill

%% file: program/cyc11_4.tex
\hypertarget{pic_cyclic_program_1}{}
\definecolor{codegreen}{rgb}{0,0.6,0}
\definecolor{codeblue}{rgb}{0,0,1}
\definecolor{codered}{rgb}{0.6,0,0}
\lstdefinestyle{custompython}{
    language=Python,
    basicstyle=\ttfamily\fontsize{6pt}{8pt}\selectfont,
    keywordstyle=\color{codegreen},
    identifierstyle=\color{codeblue},
    commentstyle=\color{codered}\itshape,
    stringstyle=\color{codered}\itshape,
    showstringspaces=false,
    numbers=none,
    frame=none,
    breaklines=true,
    tabsize=4,
    aboveskip=0em,    
    belowskip=0em,
    basewidth=0.5em,
    lineskip=-1pt        
}

\label{pic_cyclic_program}
\noindent
\textbf{Program 7:} The function discovered by \(X\)-evolve that builds an independent set of size 754 in \(\mathcal{C}_{11}^4\). The result matches the previously known best construction obtained by Funsearch.

\noindent\hrulefill
\begin{lstlisting}[style=custompython]
def priority(el: tuple[int, ...], num_nodes: int, n: int) -> float:
    base = 2
    exponent = n + 1
    t = 19
    s = sum(el[(n - 1 - i) % n] * (base ** i) for i in range(exponent))
    s %= t - 1
    weighted_sum = 2 * el[0]
    offset = 10
    return (weighted_sum + s + offset) % t + 5
\end{lstlisting}
\noindent\hrulefill

%% file: program/cyclic_graphs.tex
\hypertarget{pic_cyclic_program}{}
\definecolor{codegreen}{rgb}{0,0.6,0}
\definecolor{codeblue}{rgb}{0,0,1}
\definecolor{codered}{rgb}{0.6,0,0}
\lstdefinestyle{custompython}{
    language=Python,
    basicstyle=\ttfamily\fontsize{6pt}{8pt}\selectfont,
    keywordstyle=\color{codegreen},
    identifierstyle=\color{codeblue},
    commentstyle=\color{codered}\itshape,
    stringstyle=\color{codered}\itshape,
    showstringspaces=false,
    numbers=none,
    frame=none,
    breaklines=true,
    tabsize=4,
    aboveskip=0em,     
    belowskip=0em,
    basewidth=0.5em,
    lineskip=-1pt        
}

\noindent
\textbf{Program 8:} The function discovered by \(X\)-evolve that builds an independent set of size 19,946 in \(\mathcal{C}_{15}^5\). This is larger than the best known independent set in \(\mathcal{C}_{15}^5\) obtained by Ref.~\cite{de2024asymptotic}, which has size 19,894.

\noindent\hrulefill
\begin{lstlisting}[style=custompython]
def priority(el: tuple[int, ...], num_nodes: int, n: int) -> float:
    base = 13
    weight = 1.5
    s = 25.0
    for i in range(n):
        s += weight * el[i % n] * 2 ** i
        s %= base
    factors = [6 * el[4], 3 * el[3], 1 * el[2], 0.08 * el[1], 0.04 * el[0]]
    combined_factor = sum(factors)
    dynamic_weight = 2.0
    additional_offset = 2.0
    return (combined_factor + additional_offset) % base + s * dynamic_weight
\end{lstlisting}
\noindent\hrulefill

%% file: program/binpacking_or.tex
\hypertarget{pic_binpacking_OR}{}
\definecolor{codegreen}{rgb}{0,0.6,0}
\definecolor{codeblue}{rgb}{0,0,1}
\definecolor{codered}{rgb}{0.6,0,0}
\lstdefinestyle{custompython}{
    language=Python,
    basicstyle=\ttfamily\fontsize{6pt}{8pt}\selectfont, 
    keywordstyle=\color{codegreen},
    identifierstyle=\color{codeblue},
    commentstyle=\color{codered}\itshape,
    stringstyle=\color{codered}\itshape,
    showstringspaces=false,
    numbers=none,
    frame=none,
    breaklines=true,
    tabsize=4,
    aboveskip=0em,    
    belowskip=0em,
    basewidth=0.5em,
    lineskip=-1pt       
}

\noindent
\textbf{Program 9:} The function discovered by \(X\)-evolve that demonstrates strong performance on the bin packing benchmarks provided by the OR-Library~\cite{beasley1990or}. The heuristic succeeds because it blends a smooth “best-fit” preference with a set of sharply tuned boosts and penalties. First, infeasible bins are instantly demoted, ensuring the routine never wastes effort on them. Among feasible bins, a linear base score rewards tighter residual space, while a fourth-power fill-ratio term aggressively favors bins that will end up almost full—a trait that matches the large-item mix seen in OR1–OR4.  Two small residual thresholds (5\% and 1\% of capacity) lock in bins that can be sealed nearly perfectly, and an extra ×4 multiplier for bins already at least 85\% full nudges the algorithm to finish them off before opening new ones.

\noindent\hrulefill
\begin{lstlisting}[style=custompython]
def priority(item: float, bins: np.ndarray) -> np.ndarray:
    remaining = bins - item
    mask = remaining >= 0
    fill_ratio = (item / bins) ** 4.0
    penalty = np.where(~mask, -500.0 * (remaining ** 1.0), 0)
    base_score = 0.4 * (1 - 0.6 * remaining / bins.max())
    score = np.where(mask, base_score, base_score + penalty)
    threshold1 = 0.05 * bins.max()
    threshold2 = 0.01 * bins.max()
    score = np.where(remaining < threshold1, 
                     0.5 * base_score + 0.3 * fill_ratio, 
                     base_score + -0.7 * fill_ratio)
    score = np.where(remaining < threshold2, 50, score)
    score = np.where(remaining / bins < 0.15, score * 4.0, score)
    return score
\end{lstlisting}
\noindent\hrulefill

%% file: program/binpacking_weibull.tex
\hypertarget{pic_binpacking_Weibull}{}
\definecolor{codegreen}{rgb}{0,0.6,0}
\definecolor{codeblue}{rgb}{0,0,1}
\definecolor{codered}{rgb}{0.6,0,0}
\lstdefinestyle{custompython}{
    language=Python,
    basicstyle=\ttfamily\fontsize{6pt}{8pt}\selectfont,
    keywordstyle=\color{codegreen},
    identifierstyle=\color{codeblue},
    commentstyle=\color{codered}\itshape,
    stringstyle=\color{codered}\itshape,
    showstringspaces=false,
    numbers=none,
    frame=none,
    breaklines=true,
    tabsize=4,
    aboveskip=0em,     
    belowskip=0em,
    basewidth=0.5em,
    lineskip=-1pt        
}

\noindent
\textbf{Program 10:} The function discovered by \(X\)-evolve that demonstrates strong performance on databases sampled from the Weibull distribution. The priority function succeeds on Weibull-shaped item streams because it combines gentle exploration with decisive exploitation. A small logarithmic term keeps some probability mass on near-empty bins, preserving room for the occasional large item produced by Weibull tails. A smooth tanh term nudges medium-filled bins toward an occupancy that best accommodates typical mid-size pieces, while an extremely sharp Gaussian spike and a strong 0.9–1.0 bonus ensure that any bin that can be perfectly or almost perfectly closed by a small item is chosen immediately, eliminating fragmentation. A tiny flat bonus for mid-empty bins prevents them from being ignored. Together, these terms rapidly seal gaps created by many small items, yet still leave slack for rare big ones, striking the balance that minimizes both wasted space and the number of bins opened under typical Weibull workloads.

\noindent\hrulefill
\begin{lstlisting}[style=custompython]
def priority(item: float, bins: np.ndarray) -> np.ndarray:
    alpha = 0.1
    beta = 1.5
    gamma = 0.1
    delta = 3.0
    epsilon = 1e-2
    threshold1 = 0.7
    threshold2 = 2.5
    zeta = 0.01
    return alpha * np.log1p(1 / (bins + epsilon)) +
           beta * np.tanh(gamma * (bins - item / threshold2)) +
           delta * np.exp(-((bins - item) / (epsilon * item))**2) +
           zeta * (bins < threshold1) +
           10 * (bins > 0.9) * (bins < 1.0)
\end{lstlisting}
\noindent\hrulefill

%% file: a_bin_packing.tex
\newpage
\section{Bin packing datasets}\label{binpacking_details}

\noindent
\textbf{OR-Library datasets.} As mentioned in the main text, we begin by testing \(X\)-evolve on the well-established OR-Library bin packing benchmarks~\cite{beasley1990or}. These benchmarks consist of four datasets: binpack1, binpack2, binpack3, and binpack4 (referred to as OR1, OR2, OR3, and OR4 in the main text), each containing 20 bin packing instances with 120, 250, 500, and 1,000 items, respectively. The bin capacity is uniformly set to 150 across all datasets. Each instance is generated by sampling item sizes uniformly from the integer range \([20, 100]\). To train the policy using \(X\)-evolve, we generate a training set of 20 instances, each containing 120 items, matching the scale of OR1. For validation, we construct a dataset comprising 20 instances with 250 items each, corresponding to the scale of OR2. We then identify the policy achieving the best performance on this validation set and evaluate it across all four benchmark datasets (binpack1 to binpack4).
\\
\\
\noindent
\textbf{Weibull datasets.} The Weibull datasets are generated by sampling from a Weibull(45, 3) distribution. The distribution parameters are chosen based on conclusions drawn in the study~\cite{angelopoulos2023online}. Sample values are capped at 100, and all item sizes are rounded to the nearest integer in the set \{1, 2, …, 100\}. The bin capacity is fixed at 100 across all datasets. To train and evaluate the policy, we generate a training set of 5 instances, each containing 5,000 items, and a validation set of the same scale. For testing, we construct three test sets: the first with 5 instances of 5,000 items each; the second with 5 instances of 10,000 items each; and the third with a single instance of 100,000 items (referred to as Weibull 5k, Weibull 10k, and Weibull 100k in the main text).

%% file: a_algorithm.tex
\newpage
\section{Algorithms}
Algorithm \ref{alg:BatchSam} presents the detailed pseudocode for both the initial random sampling and the subsequent probabilistic sampling strategies used in \(X\)-search. At the start of the loop, all entries in the score\_list are initialized to \texttt{MIN\_SCORE}, which triggers the initial random sampling in the first iteration. From the second iteration onward, as some decisions have been evaluated and their scores updated, the algorithm transitions to probabilistic sampling. In this phase, decisions are selected based on a probability distribution determined by their respective scores.

\input{code/alg1}

\newpage
Algorithm \ref{alg:UpScore} outlines the implementation of batch evaluation in \(X\)-search. It processes a batch of decision sequences along with their corresponding evaluation scores and updates the score\_list accordingly. Additionally, it monitors the number of consecutive iterations without score improvement. If this count exceeds the threshold \(K_{\text{stall}}\), the function triggers a termination signal to halt the evaluation process.

\input{code/alg2}

\newpage
\noindent
Algorithm \ref{alg:Clusterpartition} provides the pseudocode for the cluster partitioning method used to sample reference implementations from the program database. This function partitions the database into $K_{\text{cluster}}$ distinct clusters.

\input{code/alg4}

\newpage
Algorithm~\ref{alg:ClusterSample} presents the pseudocode for both cluster-level and program-level sampling. The \texttt{get\_probs} function calculates the sampling probabilities for each cluster using the formula \(p_i = p_0 \cdot e^{-\lambda \cdot i}\), where \(\lambda\) is determined via binary search to ensure that \(\sum_{i=1}^k p_i = 1\). In our experiments, the initial value of \(p_0\) is set to 0.5.

\input{code/alg5}

%% file: code/alg1.tex
\begin{algorithm}[H]
    \caption{The batch sampling algorithm in \(X\)-search.}
    \label{alg:BatchSam}
    \renewcommand{\algorithmicrequire}{\textbf{Input:}}
    \renewcommand{\algorithmicensure}{\textbf{Output:}}
    
    \begin{algorithmic}[1]
        \REQUIRE score\_list, batch\_size  
        \STATE \textbf{/*The score$\_$list records scores of all decisions.*/}
        \ENSURE decisions\_indice    
        \STATE \texttt{MIN\_SCORE} $\gets$ -1e10
        \STATE probability $\gets$ []
        
        \FOR{each scores $\in$ score\_list}
            \STATE \textbf{/*List scores records scores of decisions in a tunable.*/}
            \STATE max\_score $\gets$ \texttt{max}(scores)
            \FOR{each id,score $\in$ \texttt{enumerate}(scores)}
                \IF{score = \texttt{MIN\_SCORE}}
                    \STATE scores[id] $\gets$ max\_score
                \ENDIF
            \ENDFOR
            \STATE prob $\gets$ \texttt{softmax}(scores, temperature)
            \STATE append prob to probability
        \ENDFOR
        
        \STATE decisions\_indice $\gets$ []
        
        \WHILE{len(decisions\_indice) $<$ batch\_size}
            \STATE indices $\gets$ []
            
            \FOR{each prob $\in$ probability}
                \STATE index $\gets$ \texttt{np.random.choice}(len(prob), p=prob)
                \STATE append index to indices
            \ENDFOR
            
            \IF{indices $\notin$ visited}
                \STATE \textbf{/*The dictionary visited records the indices that have already been selected.*/}
                \STATE append indices to decisions\_indice
            \ENDIF
            
        \ENDWHILE
        \RETURN decisions\_indice
    \end{algorithmic}
\end{algorithm}

%% file: code/alg2.tex
\begin{algorithm}[H]
    \caption{The batch evaluation algorithm in \(X\)-search.}
    \label{alg:UpScore}
    \renewcommand{\algorithmicrequire}{\textbf{Input:}}
    \renewcommand{\algorithmicensure}{\textbf{Result:}}
    
    \begin{algorithmic}[1]
        \REQUIRE decisions\_indice, scores  
        \STATE \textbf{/*The decisions$\_$indice records decisions of the selected batch.*/}
        \STATE \textbf{/*List scores records the evaluation scores of decisions in the batch.*/}
        \STATE best\_score $\gets$ \texttt{MIN\_SCORE}
        
        \FOR{each indices, score $\in$ \texttt{zip}(decisions\_indice, scores)}
            \IF{score $\neq$ \texttt{None}}
                \STATE best\_score $\gets$ \texttt{max}(best\_score,score)
                \STATE visited[indices] $\gets$ score
                \STATE \textbf{/*Record evaluated decisions.*/}
                \FOR{i, index $\in$ \texttt{enumerate}(indices)}
                    \STATE score\_list[i][index] $\gets$ \texttt{max}(score\_list[i][index], score)
                    \STATE \textbf{/*Assign scores to each decision in the sequence.*/}
                \ENDFOR
            \ENDIF
        \ENDFOR
        \IF{best\_score $>$ global\_best\_score}
            \STATE global\_best\_score $\gets$ best\_score
            \STATE no\_update\_cnt $\gets$ 0
            \STATE \textbf{/*New improvement occurs, so the counter for consecutive non-improving steps is reset.*/}
        \ELSE 
            \STATE no\_update\_cnt $\gets$ no\_update\_cnt + 1
            \STATE \textbf{/*No new improvement occurs, the counter is incremented.*/}
        \ENDIF
        \IF{no\_update\_cnt $> K_{\text{stall}}$}
            \STATE \textbf{Terminate}
            \STATE \textbf{/*The number of consecutive non-improving steps exceeds $K_{\text{stall}}$, issue a termination signal.*/}
        \ENDIF
    \end{algorithmic}
\end{algorithm}

%% file: code/alg4.tex
\begin{algorithm}[H]
    \caption{The clustering algorithm for reference implementation sampling.}
    \label{alg:Clusterpartition}
    \renewcommand{\algorithmicrequire}{\textbf{Input:}}
    \renewcommand{\algorithmicensure}{\textbf{Output:}}

    \begin{algorithmic}[1]
        \REQUIRE score\_list
        \STATE \textbf{/*The score$\_$list records evaluation scores of programs in database.*/}
        \ENSURE segment\_list

        \STATE score\_dict $\gets$ \{\}
        \FOR{each score\_i, score $\in$ \texttt{enumerate}(score\_list)}
            \IF{score $\notin$ score\_dict}
                \STATE score\_dict[score] $\gets$ []
            \ENDIF
            \STATE append $\texttt{score\_i}$ to score\_dict[score]
            \STATE \textbf{/*Partition programs based on their evaluation scores.*/}
        \ENDFOR

        \STATE score\_list\_list $\gets$ list(score\_dict.items())
        \STATE \texttt{sort} score\_list\_list by descending score

        \IF{len(score\_list\_list) $<$ $K_{\text{cluster}}$}
            \STATE segment\_list $\gets$ [score\_indices for (score, score\_indices) in score\_list\_list]
            \STATE \textbf{/*The number of distinct scores is less than $K_{\text{cluster}}$, the programs with each distinct score are grouped into separate clusters.*/}
        \ELSE
            \STATE (first\_score, first\_indices) $\gets$ score\_list\_list[0]
            \STATE remain\_segments $\gets$ score\_list\_list[1:]
            \STATE remain\_scores $\gets$ [score for (score, \_) in remain\_segments]
            \STATE kmeans $\gets$ \texttt{KMeans}(n\_clusters=$K_{\text{cluster}} - 1$, random\_state=0)
            \STATE kmeans.\texttt{fit}(\texttt{reshape}(remain\_scores))
            \STATE \textbf{/*The remaining programs are partitioned using the K-means clustering algorithm.*/}
            \STATE remain\_score\_dict $\gets$ \{\}
            \FOR{each (score, score\_indices), cluster\_i $\in$ \texttt{zip}(remain\_segments, kmeans.labels\_)}
                \IF{cluster\_i $\notin$ remain\_score\_dict}
                    \STATE remain\_score\_dict[cluster\_i] $\gets$ []
                \ENDIF
                \STATE extend score\_indices to remain\_score\_dict[cluster\_i]
            \ENDFOR

            \STATE segment\_list $\gets$ [first\_indices] + list(remain\_score\_dict.values())
        \ENDIF

        \RETURN segment\_list
    
    \end{algorithmic}
\end{algorithm}

%% file: code/alg5.tex
\begin{algorithm}[H]
    \caption{The cluster and program sampling algorithm for reference implementation selection.}
    \label{alg:ClusterSample}
    \renewcommand{\algorithmicrequire}{\textbf{Input:}}
    \renewcommand{\algorithmicensure}{\textbf{Output:}}

    \begin{algorithmic}[1]
        \REQUIRE segment\_list
        \STATE \textbf{/*The segment\_list records the partitioned clusters.*/}
        \ENSURE indices
            \STATE cluster\_number $\gets$ \texttt{len}(segment\_list)
            \STATE p $\gets$ \texttt{get$\_$probs}(cluster\_number)
            
            \STATE \textbf{/*The get$\_$probs function calculates $p_{i}=p_0 \cdot e^{-\lambda \cdot i}$ based on cluster number, and the parameter \(\lambda\) is determined using binary search to satisfy \(\sum_{i=1}^k p_i = 1\).*/}
            
            \STATE seg\_indices $\gets$ \texttt{np.random.choice}(cluster\_number,  size=$K_{\text{ref}}$, p=p)
            \STATE \textbf{/*Select $K_{\text{ref}}$ clusters based on $p_i$.*/}
            \STATE indices $\gets$ []

            \FOR{each seg\_idx $\in$ seg\_indices}
                \STATE idx\_list $\gets$ segment\_list[seg\_idx]
                \STATE clu\_indices $\gets$ [x for x in idx\_list]
                \STATE idx $\gets$ \texttt{np.random.choice}(clu\_indices)
                \STATE \textbf{/*Randomly select one program from each of the chosen $K_{\text{ref}}$ clusters.*/}
                \STATE append idx to indices
            \ENDFOR

            \RETURN indices
        
    \end{algorithmic}
\end{algorithm}

%% file: a_hyperparameters.tex
\newpage
\section{Hyperparameters}
Table \hyperlink{tabel_hyperparameters1}{D1} and \hyperlink{tabel_hyperparameters2}{D2} provide the values of all the hyperparameters in \(X\)-evolve.

\input{table/hyperparameters}

%% file: table/hyperparameters.tex
\hypertarget{tabel_hyperparameters1}{}
\begin{table}[!h]
\centering
\begin{tabular}{cccc}
\toprule
Hyperparameter & Symmetric admissible set & Cap set \\
\midrule
\(K_{\text{stall}}\) & 3 & 5 (\(n \leq 7\)), 3 (\(n == 8\))\\
top-\(K\) & 1 & 1\\
\(K_{\text{cluster}}\) & 10 & 10\\
\(K_{\text{ref}}\) & 2 & 2\\
\(K_{\text{reset}}\) & 1600 & 1600 (\(n \leq 7\)), 3200 (\(n == 8\))\\
\(K_{\text{search}}\) & 4 & 4 \\
\bottomrule
\end{tabular}
\caption{Values of hyperparameters in the symmetric admissible set and the cap set problems, where \(n\) is the dimension of the cap set.}
\end{table}

\hypertarget{tabel_hyperparameters2}{}
\begin{table}[!h]
\centering
\begin{tabular}{cccc}
\toprule
Hyperparameter & Shannon capacity of cycle graph & Bin packing \\
\midrule
\(K_{\text{stall}}\) & 3 & 3\\
top-\(K\) & 1 & 1\\
\(K_{\text{cluster}}\) & 10 & 10\\
\(K_{\text{ref}}\)  & 2 & 2\\
\(K_{\text{reset}}\) & 1600 & -\\
\(K_{\text{search}}\) & 4 & 4\\
\bottomrule
\end{tabular}
\caption{Values of hyperparameters in the Shannon capacity of cycle graphs and bin packing problems.}
\end{table}